\documentclass[conference]{IEEEtran}
\usepackage{fancyhdr}
\usepackage[a4paper, left=0.6in, right=0.6in, top=0.6in, bottom=1in]{geometry}

\usepackage[utf8]{inputenc}    
\usepackage[T1]{fontenc}       
\usepackage[english]{babel}    

\usepackage{inconsolata}       
\usepackage{microtype}         

\usepackage{graphicx}          
\usepackage{subcaption}        
\usepackage{booktabs}          
\usepackage{multirow}          
\usepackage{tabularx}          
\usepackage{xcolor}            
\definecolor{lightgreen}{RGB}{144,238,144}
\definecolor{target}{HTML}{32CD32}
\definecolor{unpredicted}{HTML}{F28C8C}
\definecolor{candidate}{HTML}{87CEFA}
\definecolor{history}{HTML}{D3D3D3}

\usepackage{amsmath,amssymb,amsfonts} 

\usepackage{cite}              
\usepackage{url}               

\usepackage{enumitem}

\usepackage[most]{tcolorbox}

\usepackage{fontsize}

\date{}
\title{\fontsize{22pt}{22pt}\selectfont ContrastiveGaussian: High-Fidelity 3D Generation with Contrastive Learning and Gaussian Splatting}

\author{
Junbang Liu\textsuperscript{1,*}, Enpei Huang\textsuperscript{1,*}, Dongxing Mao\textsuperscript{2}, Hui Zhang\textsuperscript{1,$\dagger$}, Xinyuan Song\textsuperscript{3}, Yongxin Ni\textsuperscript{2} \vspace{0.15cm}\\
\textsuperscript{1}Beijing Normal-Hong Kong Baptist University, Zhuhai, China \\
\textsuperscript{2}National University of Singapore, Singapore, Singapore \\ \textsuperscript{3}Emory University, Atlanta, USA \\
r130026088@mail.uic.edu.cn \qquad r130026051@mail.uic.edu.cn \qquad e0724629@u.nus.edu \\
amyzhang@uic.edu.cn \qquad\qquad xsong30@emory.edu \qquad niyongxin@u.nus.edu
}

\begin{document}

\maketitle

\thispagestyle{fancy}

\fancyhf{}
\renewcommand{\headrulewidth}{0pt}
\renewcommand{\footrulewidth}{0pt}

\renewcommand{\footnoterule}{%
  \vspace{0.1cm} 
  \hrule width 4.4cm
  \vspace{0.1cm}
}
\footnotetext[1]{Equal Contribution.}
\footnotetext[2]{Corresponding Author.}
\footnotetext{This work is supported in part by the Natural Science Foundation of China (62076029); in part by the National Key R$\&$D Program of China (2022YFE0201400); in part by the Guangdong Provincial Key Laboratory of IRADS (2022B1212010006) and in part by Guangdong Higher Education Upgrading Plan (2021-2025) with No. of UICR0400006-24.}


\begin{abstract}
\noindent Creating 3D content from single-view images is a challenging problem that has attracted considerable attention in recent years. Current approaches typically utilize score distillation sampling (SDS) from pre-trained 2D diffusion models to generate multi-view 3D representations. Although some methods have made notable progress by balancing generation speed and model quality, their performance is often limited by the visual inconsistencies of the diffusion model outputs. In this work, we propose ContrastiveGaussian, which integrates contrastive learning into the generative process. By using a perceptual loss, we effectively differentiate between positive and negative samples, leveraging the visual inconsistencies to improve 3D generation quality. To further enhance sample differentiation and improve contrastive learning, we incorporate a super-resolution model and introduce another Quantity-Aware Triplet Loss to address varying sample distributions during training. Our experiments demonstrate that our approach achieves superior texture fidelity and improved geometric consistency. Code will be available at https://github.com/YaNLlan-ljb/ContrastiveGaussian.
\end{abstract}

\section{Introduction}
Automated 3D content generation from a single-view image has made remarkable progress in recent years, becoming a crucial technology in many fields such as virtual reality (VR), augmented reality (AR) and digital entertainment. However, this task remains inherently challenging due to the limited information provided by single-view images, which often fail to capture the full geometric structure and texture details of a scene, resulting in ambiguities and inaccuracies of generated 3D content~\cite{1,2}.

The researches on 3D content creation can be broadly categorized into two main approaches: \textit{inference-based 3D native methods}, which directly generate 3D structures from images, and \textit{optimization-driven 2D lifting methods}, which rely on transforming 2D representations into 3D models~\cite{11}. Recently, Score Distillation Sampling (SDS) introduced by DreamFusion enables advanced 3D generation by using robust 2D diffusion models~\cite{4} to construct 3D geometries and appearances. This innovation inspired the development of 2D lifting methods~\cite{5,6,7}. However, relying solely on SDS supervision can lead to inconsistencies and ambiguities. Neural Radiance Fields (NeRF)~\cite{8} can alleviate these issues by capturing detailed 3D features. Nonetheless, their optimization remains computationally expensive, often requiring hours of training and limiting scalability~\cite{9}.

Recent methods like DreamGaussian~\cite{10} simplify optimization and accelerate generation by leveraging Gaussian splatting~\cite{3} and diffusion models. However, optimization-based approaches still suffer from visual inconsistencies~\cite{11}, such as mismatched geometry (e.g., misaligned surfaces or irregular shapes) and distorted textures characterized by noise and artifacts~\cite{3,12}. These issues significantly hinder the visual quality required for practical, high-fidelity 3D content generation.

In this paper, we introduce a novel image-to-3D framework, \textit{ContrastiveGaussian}, designed to significantly improve the fidelity and consistency of generated 3D content. Inspired by DreamGaussian, our method utilizes SDS loss to establish a robust Gaussian representation. We then integrate contrastive learning with 2D diffusion priors to refine the 3D Gaussian splatting process, leading to improved texture and geometric details. To further improve input quality, we apply a super-resolution technique to enhance the edges and details of single-view images, creating high-quality samples for contrastive learning. To address sample quality inconsistencies, we propose the \textit{Quantity-Aware Triplet Loss}, ensuring stable and effective contrastive learning despite sample imbalances. Together, these improvements significantly boost detail fidelity and visual consistency in the final 3D models. Compared to existing methods, our framework generates realistic 3D models with refined textures and geometry from a single-view image in approximately 80 seconds. \noindent Overall, our contributions are as follows: 
\begin{itemize}
    \setlength{\itemsep}{0pt} 
    \item We introduce a contrastive learning strategy into the 3D content generation field, enabling the model to produce more robust Gaussian representations.
    \item We enhance the input images by introducing a super-resolution module, substantially improving detail and texture representation during generation.
    \item We propose a Quantity-Aware Triplet Loss to address varying sample distributions, thereby improving learning efficiency and overall generation quality.
    \item Through extensive empirical evaluations, we demonstrate that ContrastiveGaussian significantly outperforms most existing methods in terms of generation speed, detail fidelity, and visual consistency.
\end{itemize}

\setlength{\textfloatsep}{5pt}    
\setlength{\floatsep}{5pt}        
\setlength{\dbltextfloatsep}{5pt} 
\setlength{\dblfloatsep}{5pt}     
\begin{figure*}[!ht]
    \centering
    \includegraphics[width=0.9\textwidth]{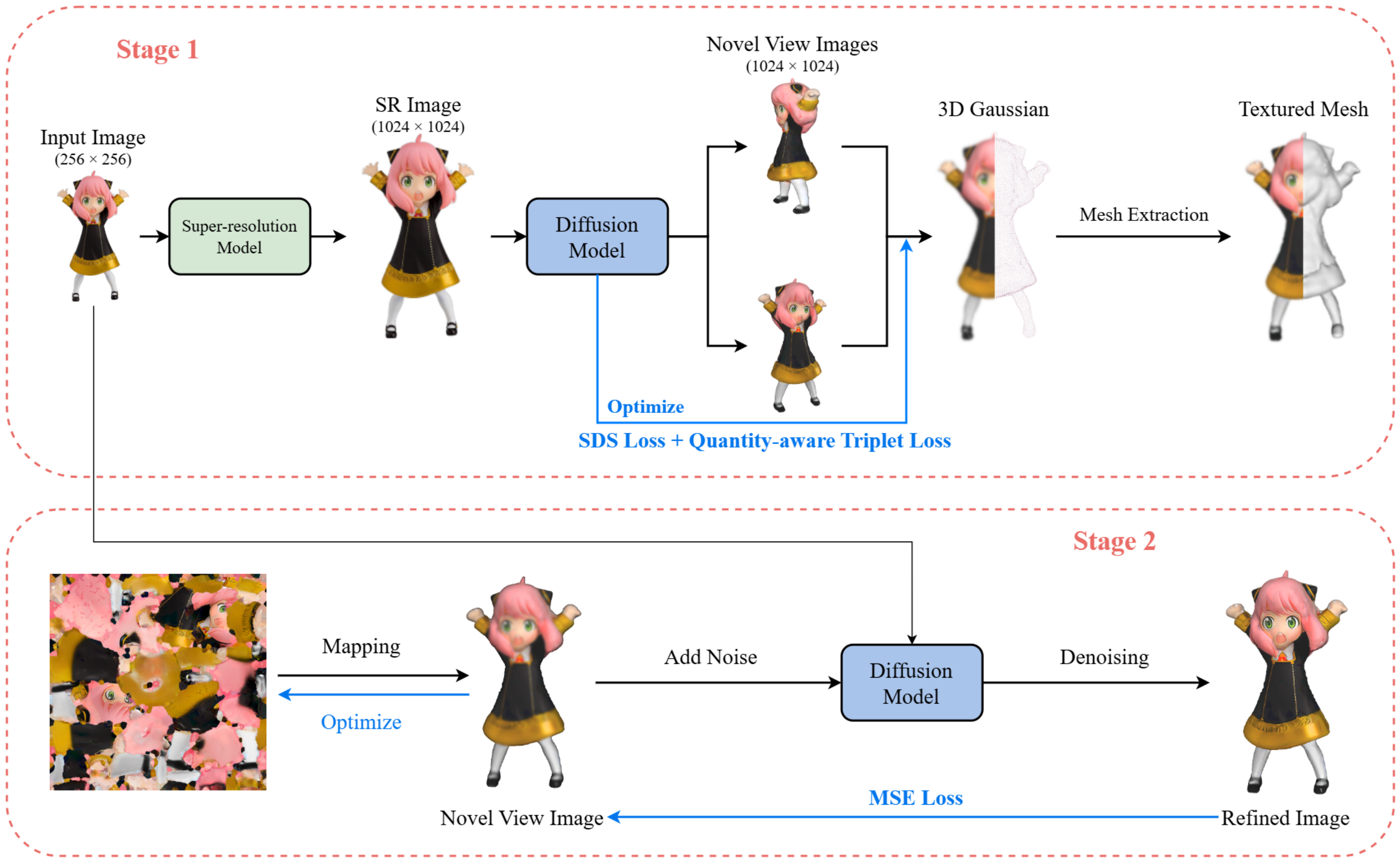}
    \caption{\textbf{ContrastiveGaussian Framework.} There are two stages in our framework. In Stage 1, the input image is upscaled using a super-resolution model, followed by optimization of the 3D Gaussian representation through SDS loss and the Quantity-Aware Triplet Loss. After obtaining refined 3D Gaussian representation, we then convert it into a textured mesh. In Stage 2, the texture details of the generated mesh are further enhanced through the application of MSE loss.}
    \label{fig:framework}
\end{figure*}

\begin{figure}[!ht]
\centering
\includegraphics[width=0.48\textwidth]{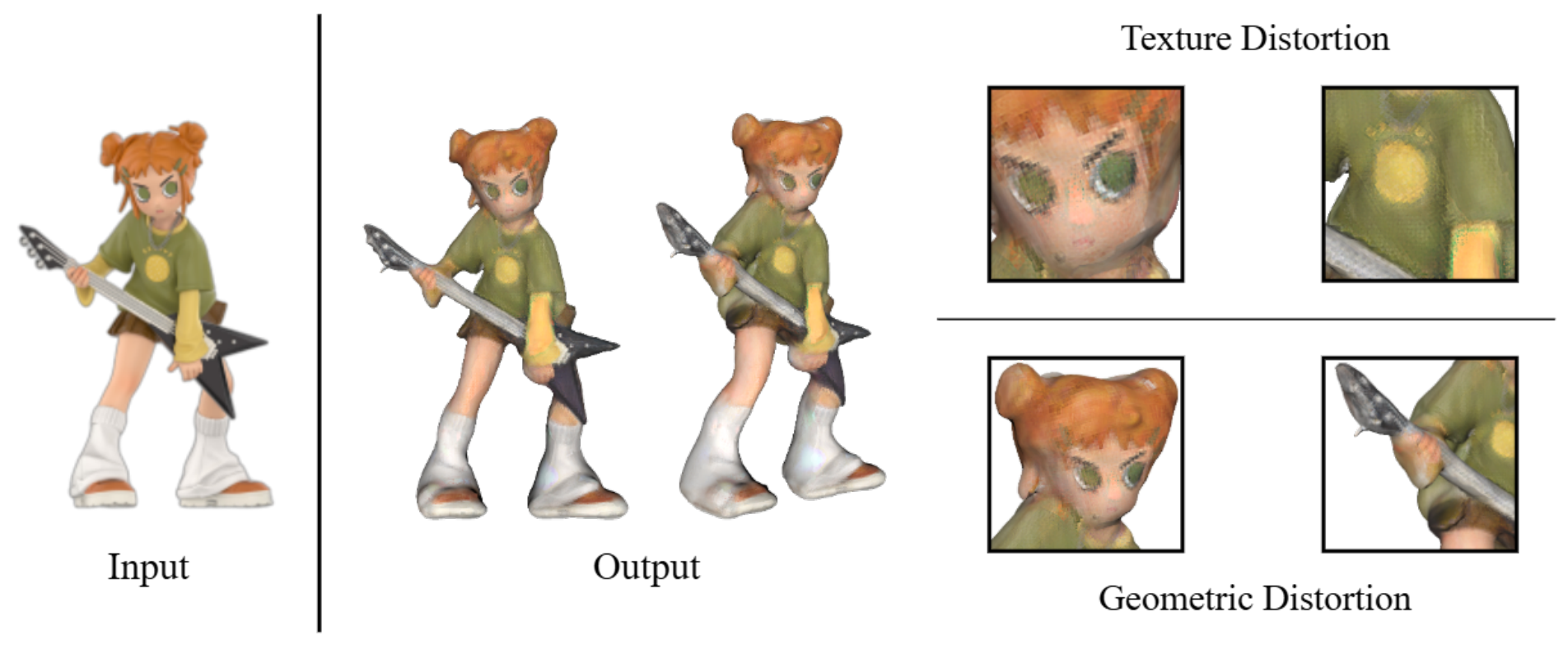}
\caption{
    \textbf{Distortion Artifacts.}
    Distortion artifacts can cause irregularities in both texture and geometry, as illustrated in this example.
}
\label{fig:distortion}
\end{figure}

\section{RELATED WORKS}

\subsection{3D Representations}
A variety of 3D representations have demonstrated promising results in image-to-3D tasks. Neural Radiant Field (NeRF) uses volumetric rendering and positional encoding to perform 3D optimization based solely on 2D supervision. This approach has made NeRF widely adopted for tasks in both 3D generation~\cite{3,5} and reconstruction \cite{13,14,15,16}.  However, optimizing NeRF models is time-consuming and often requires multi-view inputs, which can be problematic. Although some efforts have attempted to accelerate NeRF training~\cite{17,18}, these works primarily focus on reconstruction rather than generation. 

Recently, 3D Gaussian splatting~\cite{19} has emerged as an alternative representation to NeRF, achieving remarkable results and faster generation speeds. By optimizing 3D Gaussian distributions and employing efficient differentiable rendering, Gaussian splatting significantly accelerates rendering and supports real-time applications without relying on spatial pruning. Despite its success in reconstruction tasks, relatively few studies have explored its potential in generative settings. To address this gap, we adapt 3D Gaussian splatting for image-to-3D generation, expanding its applicability and demonstrating its effectiveness beyond reconstruction.

\subsection{Image-to-3D Content Generation Tasks}
Image-to-3D content generation refers to generate 3D content from a single reference image, typically a front-view image. Recently, data-driven 2D diffusion models have demonstrated impressive results in image and video generation~\cite{20,21}. However, it is not easy to apply them to 3D generation because of the need to organize massive 3D datasets. DreamFusion~\cite{3} proposes Score Distillation Sampling (SDS), a method for extracting 3D content from pre-trained 2D diffusion models by rendering from multiple viewpoints. Based on the SDS, Zero-1-to-3~\cite{22} enables novel view synthesis conditioned on relative camera poses. While this approach allows to achieve higher quality 3D generation, it still suffers from long optimization times. In contrast, One-2-3-45~\cite{23} directly generates realistic 3D shapes from the images produced by Zero123. Motivated by these advancements, our two-stages framework, presents an efficient image-to-3D generation. Our method completes the generation in approximately 80 seconds while preserving high-quality textures and details, pushing the boundary toward practical and rapid 3D content creation.

\section{METHODOLOGY}
	In this section, we introduce ContrastiveGaussian, a two-stage image-to-3D framework as illustrated in Fig.~\ref{fig:framework}. We first upscale the input image using a super-resolution model, then generate a novel perspective via a 2D diffusion model. Next, we integrate contrastive learning and form a new triplet loss based on synthesized images. We subsequently perform 3D Gaussian splatting optimized by SDS, Reference and Quantity-Aware Triplet Loss. An efficient mesh extraction then produces a textured mesh from the Gaussian representation. Finally, we refine the coarse texture through a diffusion-based denoising process, with a multi-step MSE loss supervising the entire pipeline.

\subsection{Gaussian Splatting Generation with Contrastive Learning}

\textbf{3D Gaussian Splatting.} The 3D Gaussian splatting has proven effective in 3D content generation, providing both rapid inference and high-quality results~\cite{11, 19}. Formally, a Gaussian distribution is described by its center $\mathbf{x} \in \mathbb{R}^{3}$, scaling factor $\mathbf{s} \in \mathbb{R}^{3}$, and rotation quaternion $\mathbf{r} \in \mathbb{R}^{4}$. Additionally, an opacity value $\alpha \in \mathbb{R}$ and a color feature $\mathbf{c} \in \mathbb{R}^{3}$ are used for volumetric rendering. We denote all these parameters collectively as $\Theta$, within the parameters of the $i$-th Gaussian given by $\Theta_i = \{ \mathbf{x_i},\mathbf{s_i},\mathbf{r_i},\alpha_i,\mathbf{c_i} \}$. This work uses the efficient rendering implementation from Kerbl et al.~\cite{11} to optimize $\Theta$. In each optimization step, we render an RGB image $I^{p}_{RGB}$ and a transparency $I^{p}_{A}$ for a random sampled camera pose $p$. The underlying 3D Gaussians are then optimized via SDS and our QA-Triplet loss.

\textbf{Score Distillation Sampling.}
In the first stage, a reference image $\tilde{I}_{RGB}^r$ and a corresponding foreground mask $\tilde{I}_{A}^r$ are provided as inputs. We use the Zero-1-to-3 XL~\cite{22,24} as the 2D diffusion prior. The gradient of SDS loss can be formulated as:
\begin{equation}
\nabla_{\Theta} L_{\text{SDS}} = \mathbb{E}_{t,p,\epsilon} \left[ w(t) \left( \epsilon_{\phi} (I_{\text{RGB}}^p; t, \tilde{I}_{\text{RGB}}^r, \Delta p) - \epsilon \right) \frac{\partial I_{\text{RGB}}^p}{\partial \Theta} \right],
\end{equation}
where $w(t)$ is a weighting function, and $\epsilon_{\phi}(\cdot)$ is the predicted noise from the 2D diffusion prior $\phi$. The prediction depends on the rendered image $I^{p}_{RGB}$, the reference image $I_{RGB}^r$, the time step $t$, and $\Delta p$ which denotes the relative change in camera pose from the reference camera $r$. A reference loss is applied to optimize the reference view image $I^{r}_{RGB}$ and transparency $I_{A}^r$, ensuring alignment with the input:
\begin{equation}
\mathcal{L}_{\text{Ref}} = w_{\text{RGB}} \| I_{\text{RGB}}^r - \tilde{I}_{\text{RGB}}^r \|_2^2 + w_{\text{A}} \| I_{\text{A}}^r - \tilde{I}_{\text{A}}^r \|_2^2,
\end{equation}
where $w_{\text{RGB}}$ and $w_{\text{A}}$ are weights that increase linearly during training.

\textbf{Contrastive Learning.}
As illustrated in Fig.~\ref{fig:distortion}, variations in camera positions during multi-view image generation can lead to misalignment or distortion, especially at object edges and complex regions. This indicates that relying solely on the SDS loss is inadequate for effective model learning. To address this, we propose a contrastive learning-based approach aimed at enhancing output consistency and quality. Specifically, we leverage LPIPS (Learned Perceptual Image Patch Similarity)~\cite{25} to distinguish between positive samples (high-quality images generated from specific viewpoints using a 2D diffusion prior) and negative samples (lower-quality images). Images with lower LPIPS values are considered positive samples due to their high perceptual similarity, whereas those with higher LPIPS values serve as negative samples. Utilizing these pairs, we introduce a contrastive loss function (QA-Triplet loss) to encourage the model to learn more discriminative and generalizable features, thus effectively guiding the optimization of the underlying Gaussian distribution.

\textbf{Single-view Super-resolution.}
However, contrastive learning alone remains insufficient, as low-resolution images often lack distinguishable quality variations, limiting its effectiveness.o overcome this, we propose integrating a single-view super-resolution model~\cite{26} to upscale input images from from $256 \times 256$ to $1024 \times 1024$. Increasing the image resolution highlights both positive and negative attributes, amplifying differences between samples. This increased differentiation facilitates more effective contrastive learning, providing the model with a broader quality spectrum for alignment and discrimination.

\textbf{Quantity-Aware Triplet Loss.}
During training, we observe that the model occasionally generates exclusively positive or negative samples, limiting learning from these extreme cases. Considering that our method serves as a generalizable post-processing framework, we propose the Quantity-Aware Triplet Loss (QA-Triplet Loss), which dynamically supervises the sample ratio to effectively adapt across diverse scenarios, maximizing learning opportunities. Formally, it is defined as:
\begin{equation}
L_{\text{QA-Triplet}} = \max \Bigl(0,\, Q(p)\,d(a,p) - Q(n)\,d(a,n) + \alpha \Bigr),
\end{equation}
\begin{equation}
Q(\cdot) = \log_2\bigl(1 + N(\cdot)\bigr),
\end{equation}
where $p$ denotes the positive samples, $n$ denotes the negative samples, and $\alpha$ is the margin parameter. The anchor sample $a$ acts as a reference for distance comparisons. The function $N(\cdot)$ represents the number of samples, and the embedding distance is computed as $d(a,\cdot)=\|f(a)-f(\cdot)\|_2$. 
The logarithmic weighting $Q(\cdot)$ dynamically adjusts the influence of positive and negative samples based on their quantity. Specifically, if positive samples are absent ($N(p)=0$), the loss emphasizes enlarging the anchor-negative distance; conversely, if negative samples are absent ($N(n)=0$), it prioritizes reducing the anchor-positive distance. Additionally, when sample distances are insufficiently discriminative (below the margin $\alpha$), the loss in (3) imposes a penalty scaled by sample quantity.

The final loss is the sum of the above three losses (SDS Loss, Reference Loss, and QA-Triplet Loss), which is minimized during training to optimize the underlying Gaussian distribution parameter $\Theta$.
\begin{figure*}[t]
\centering
\setlength{\tabcolsep}{1pt}
\renewcommand{\arraystretch}{0.0}
\resizebox{0.9\textwidth}{!}{%
\begin{tabular}{c|c|c|c|cccc}

\includegraphics[width=0.11\textwidth, height=3cm]{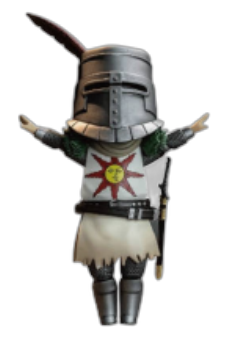} &
\includegraphics[width=0.11\textwidth, height=3cm]{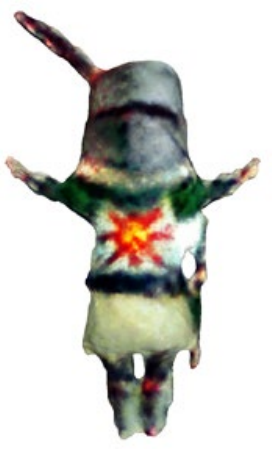} &
\includegraphics[width=0.11\textwidth, height=3cm]{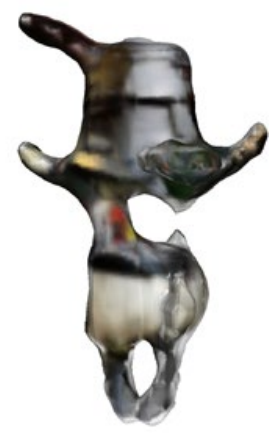} &
\includegraphics[width=0.11\textwidth, height=3cm]{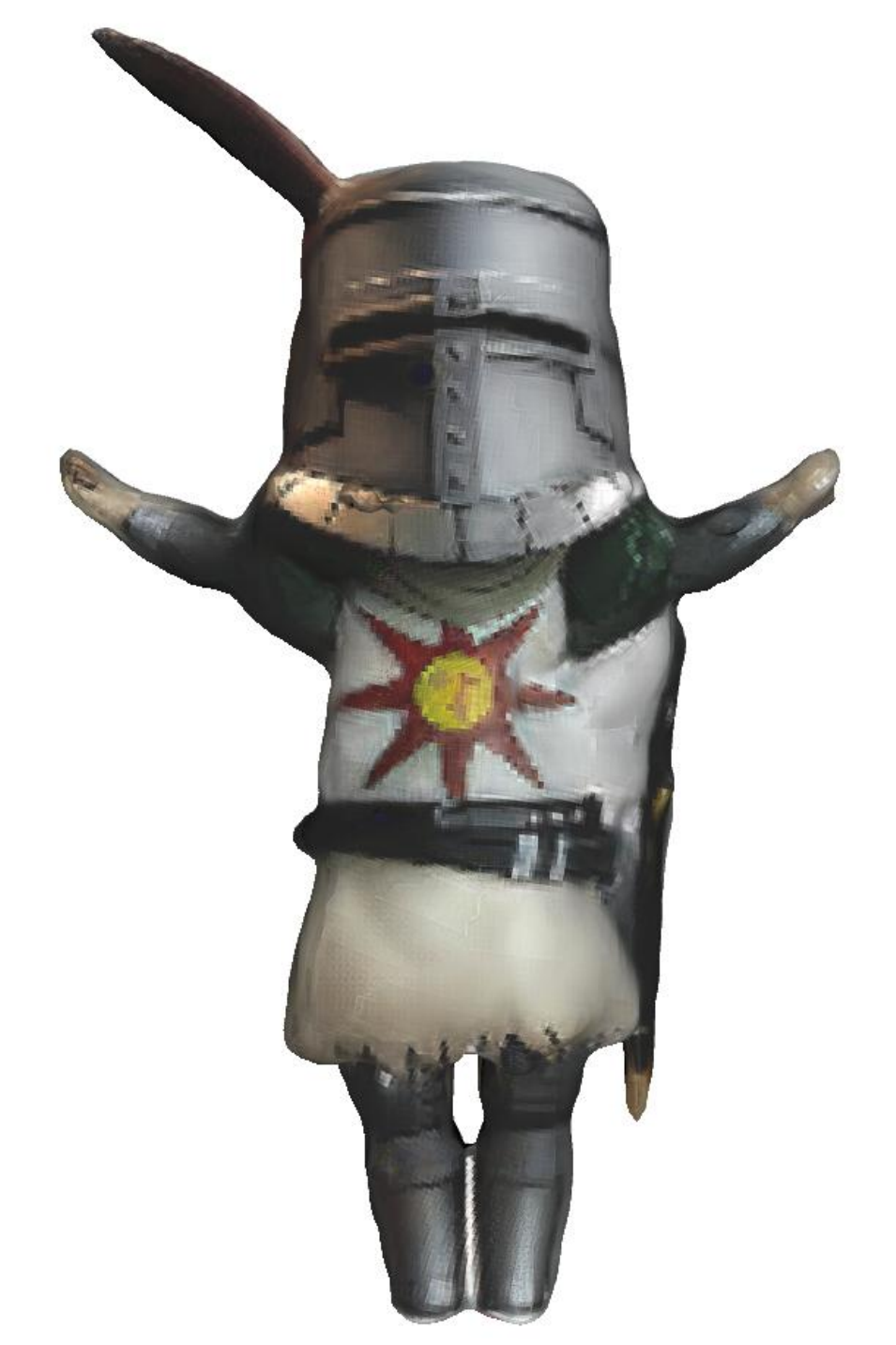} &
\includegraphics[width=0.11\textwidth, height=3cm]{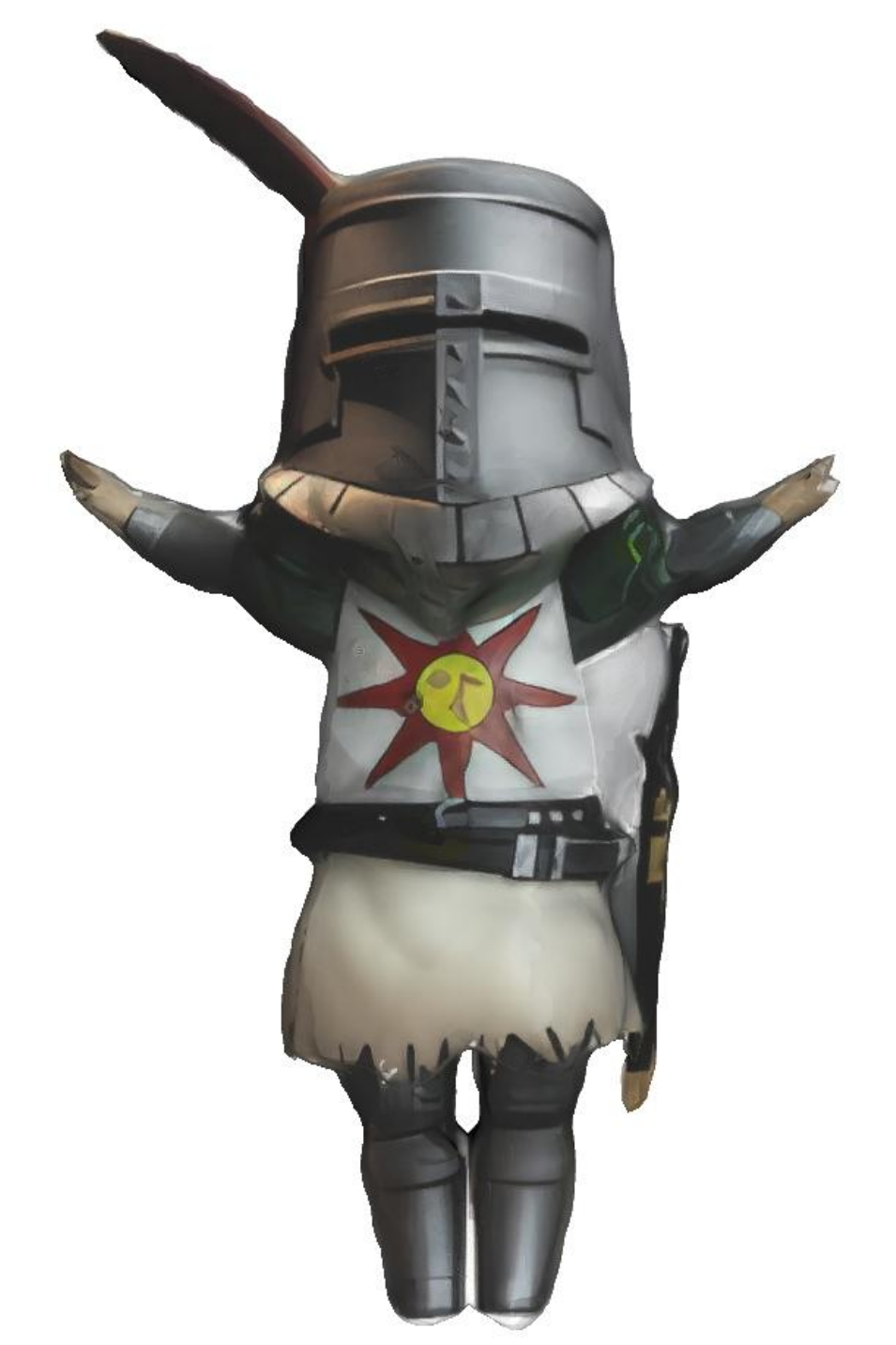} &
\includegraphics[width=0.095\textwidth]{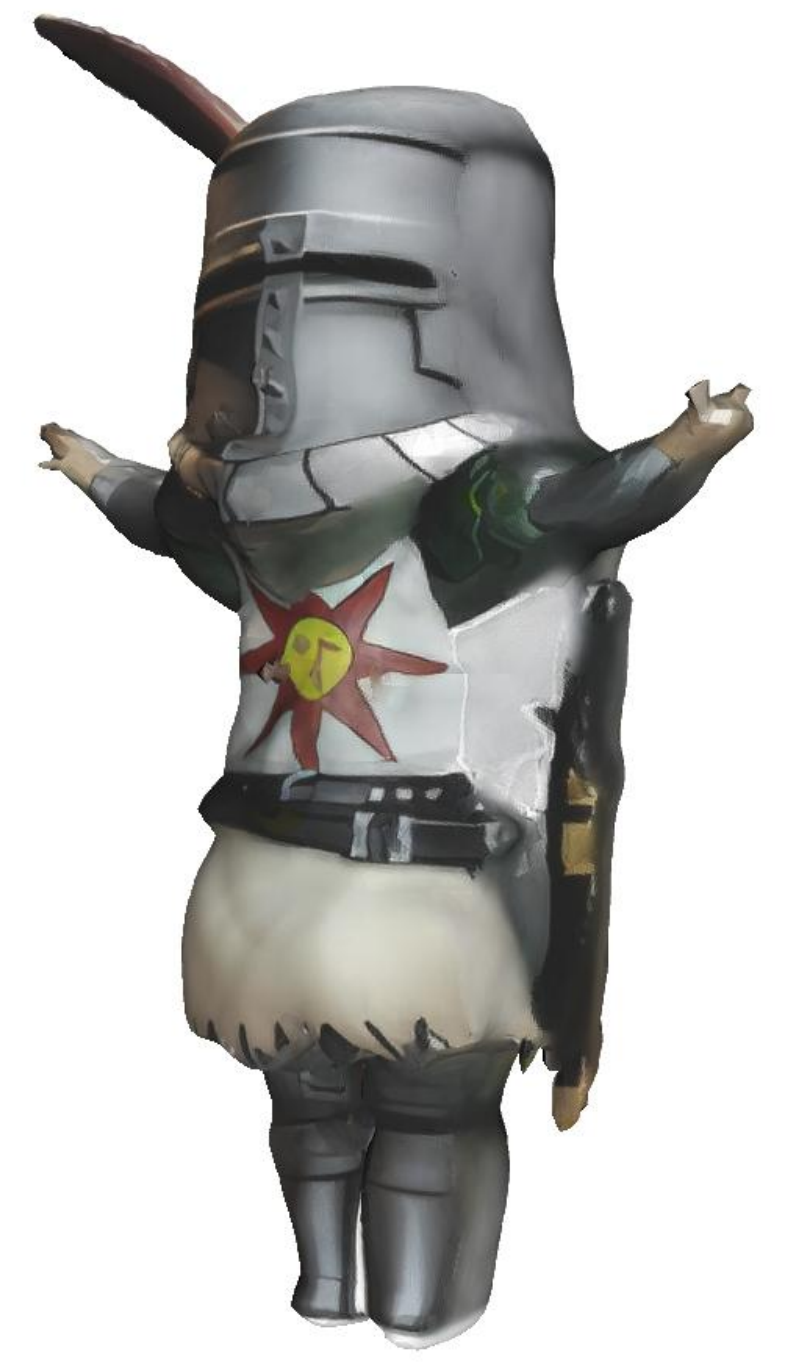} &
\includegraphics[width=0.095\textwidth]{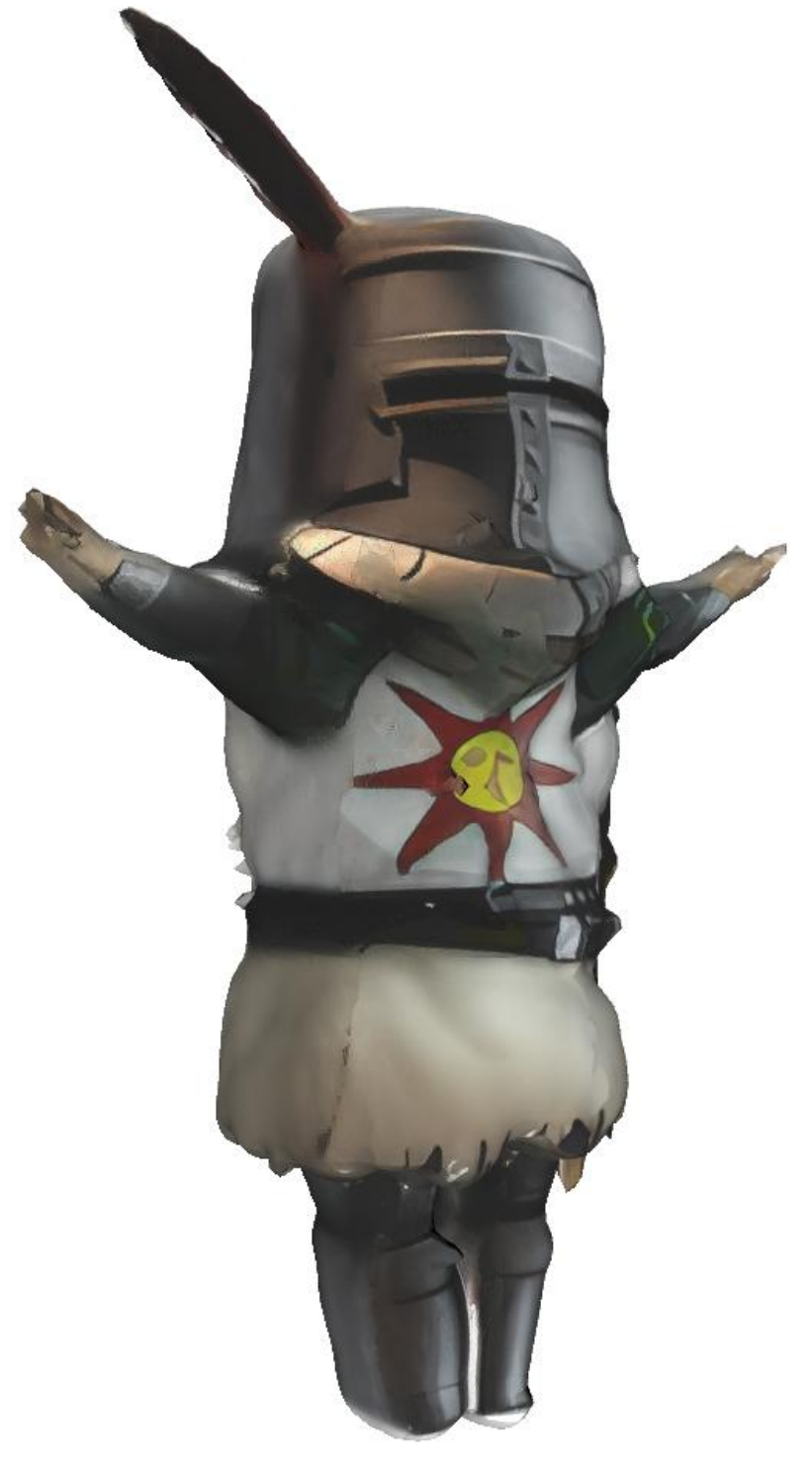} &
\includegraphics[width=0.11\textwidth, height=3cm]{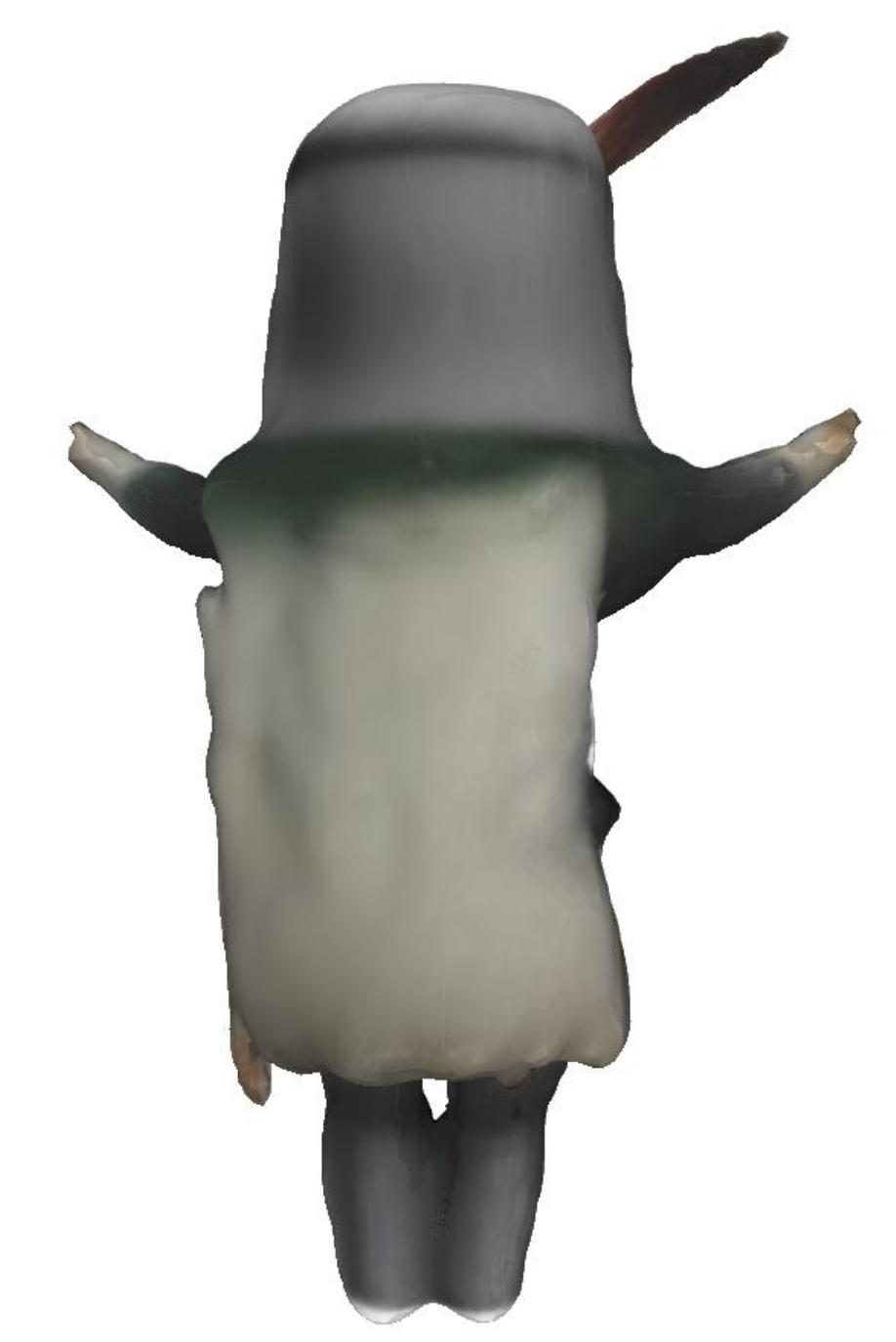} \\[-2pt]

\includegraphics[width=0.11\textwidth]{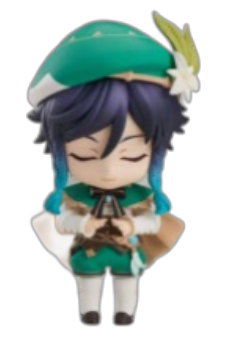} &
\includegraphics[width=0.10\textwidth, height=3cm]{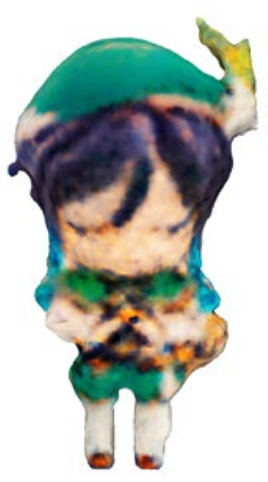} &
\includegraphics[width=1.6cm, height=2.85cm]{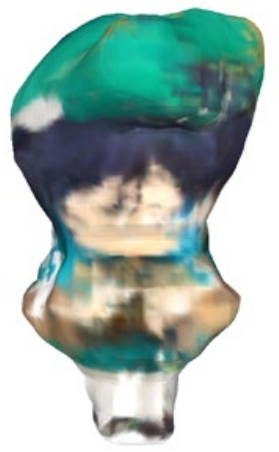} &
\includegraphics[width=0.11\textwidth]{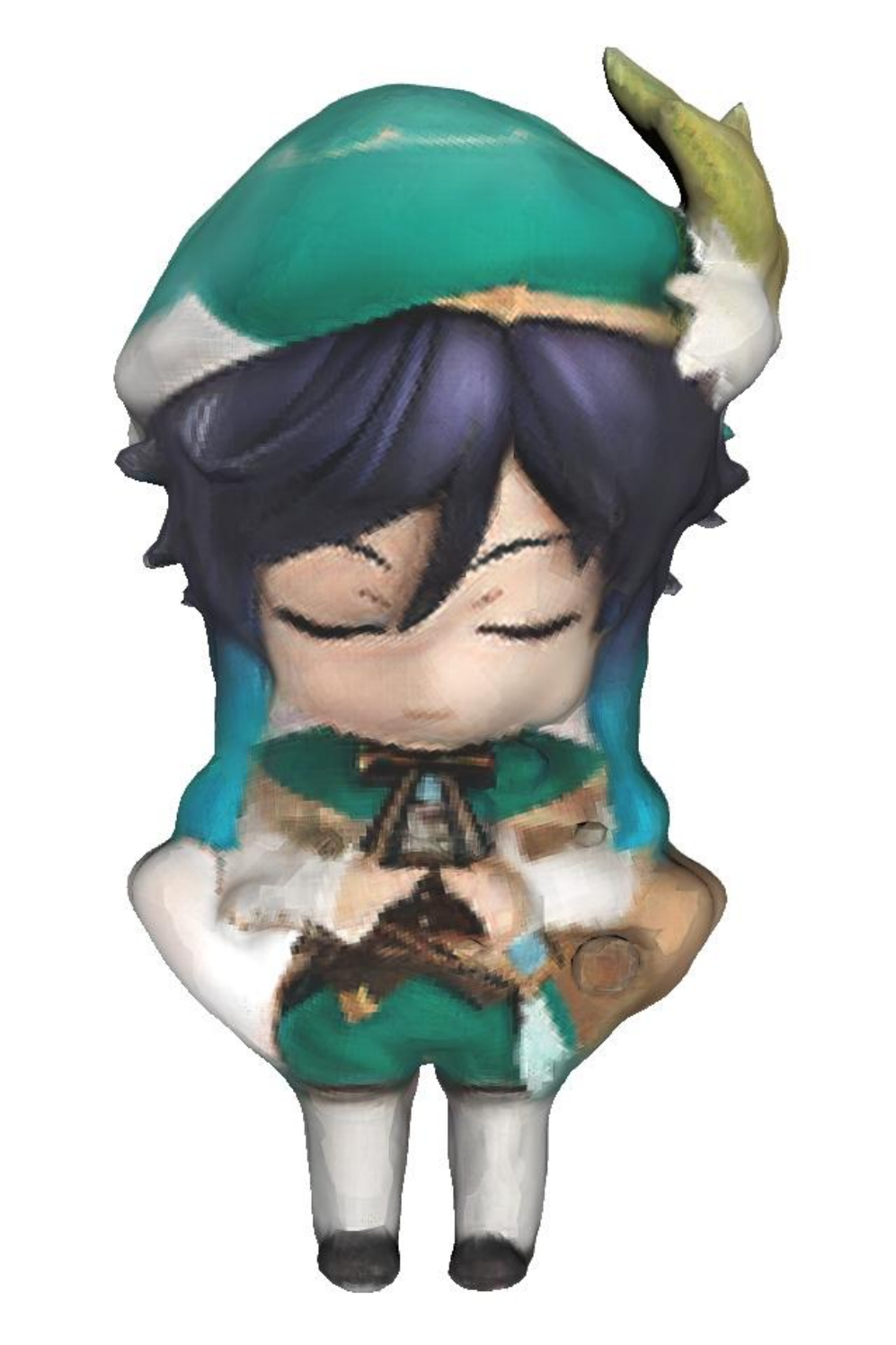} &
\includegraphics[width=0.11\textwidth]{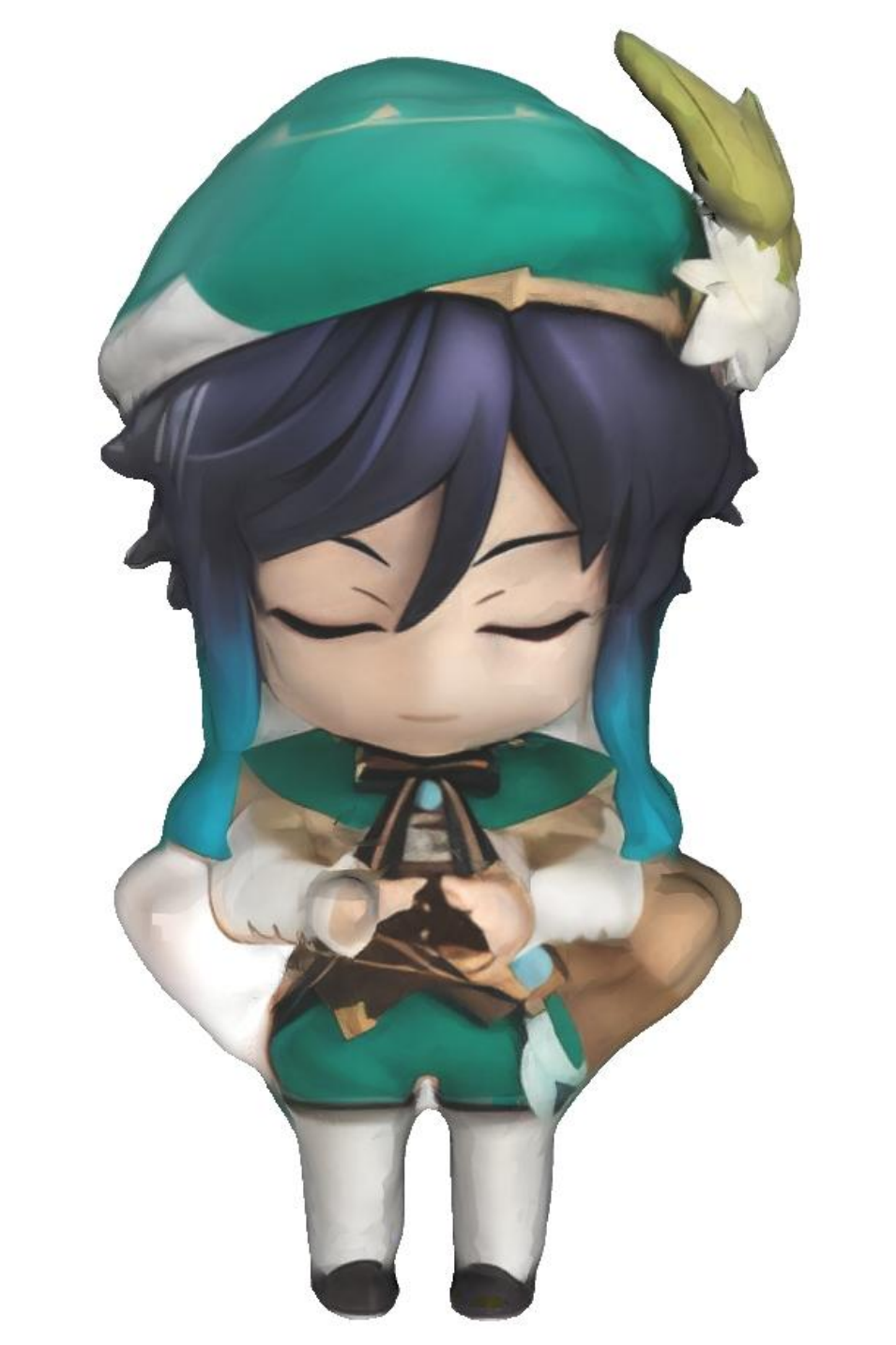} &
\includegraphics[width=0.089\textwidth]{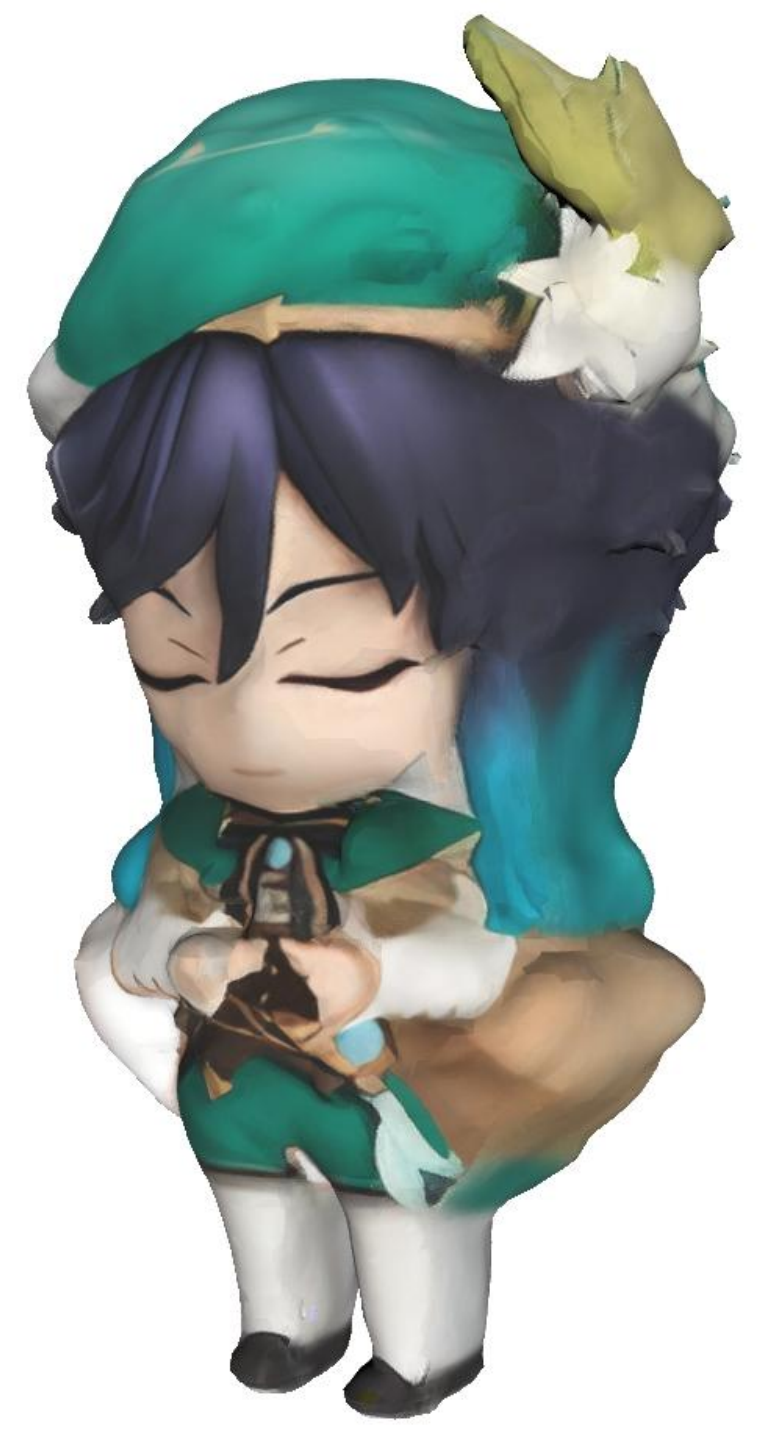} &
\includegraphics[width=0.090\textwidth]{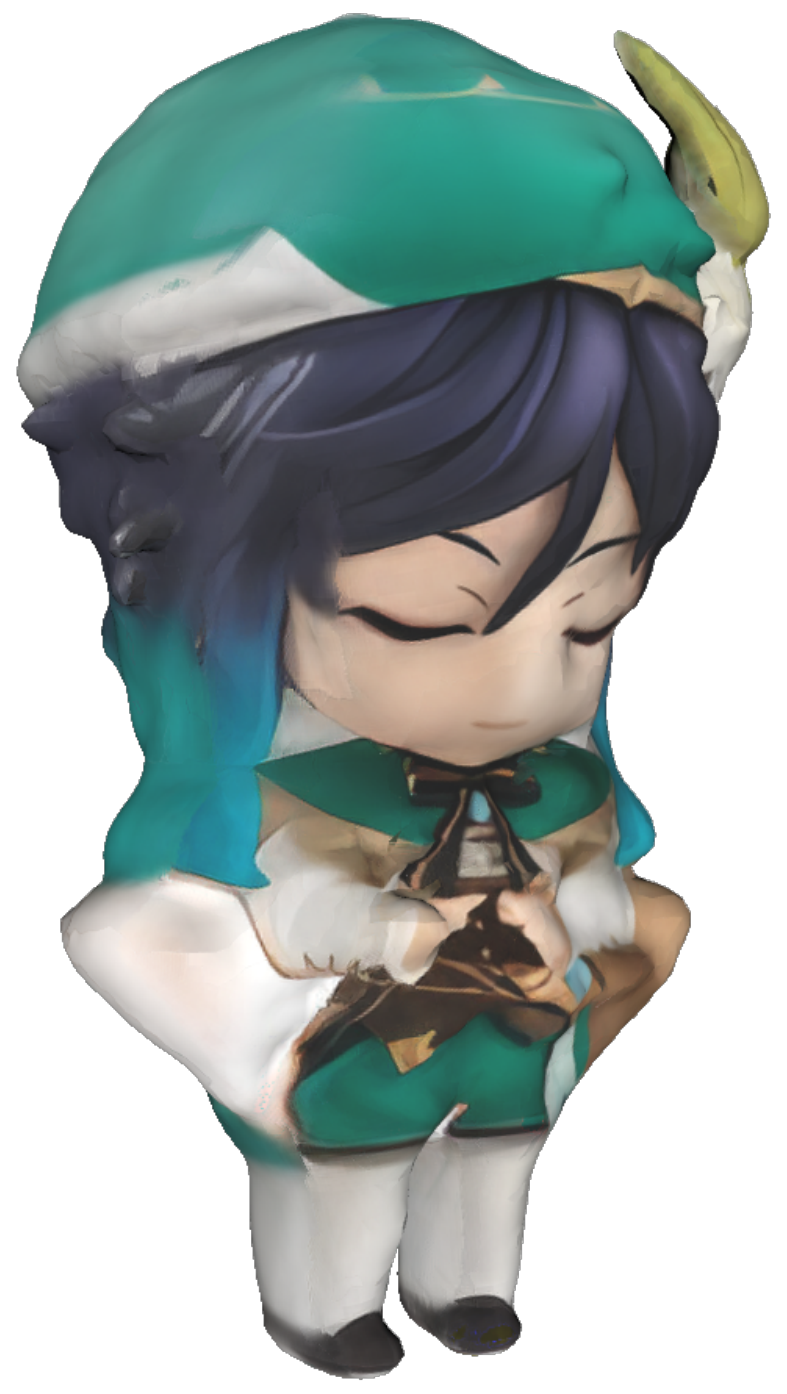} &
\includegraphics[width=0.11\textwidth]{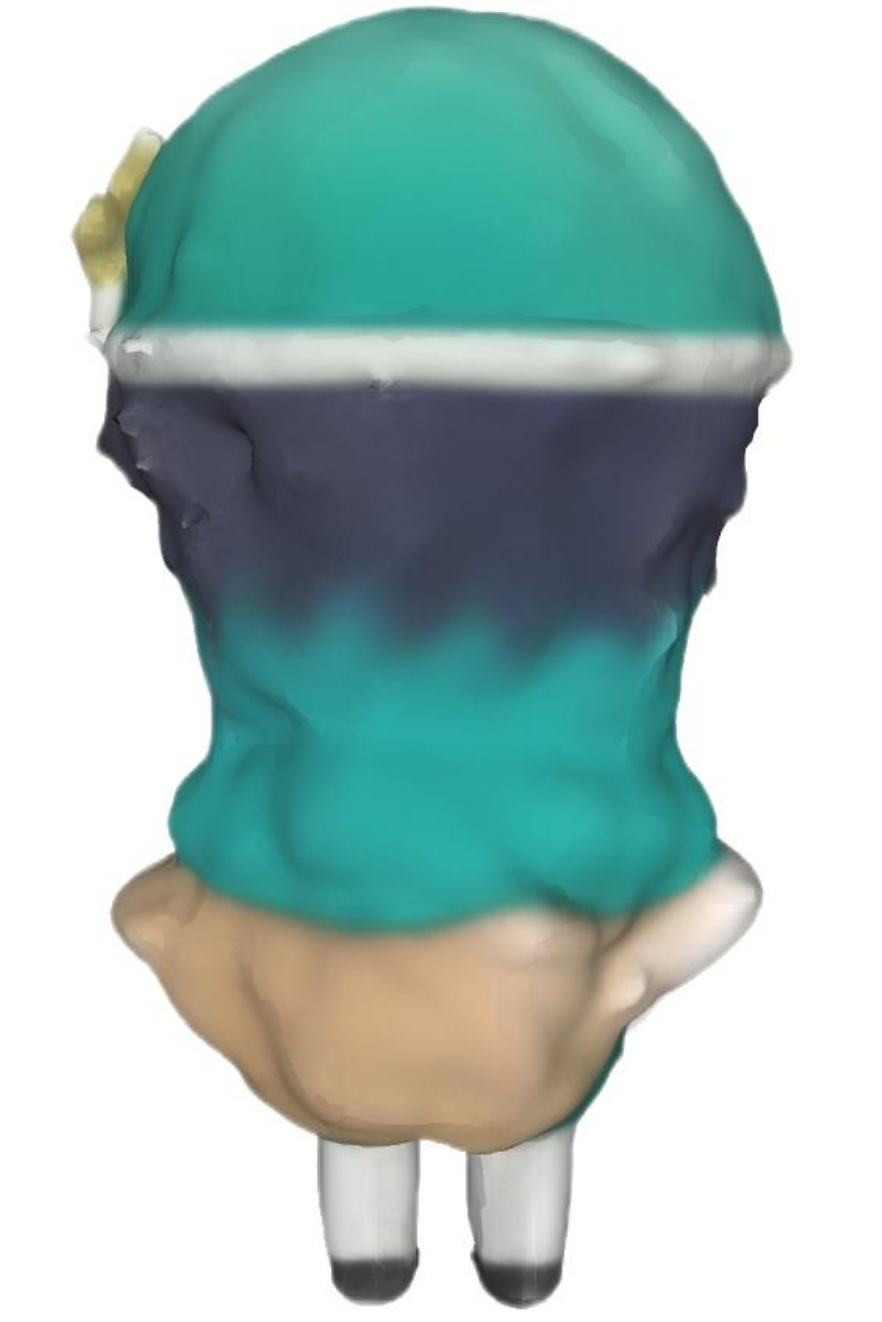} \\[-2pt]

\includegraphics[width=0.11\textwidth]{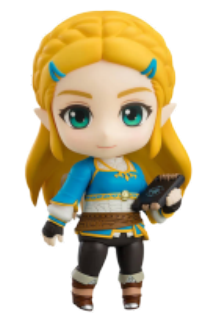} &
\includegraphics[width=0.093\textwidth]{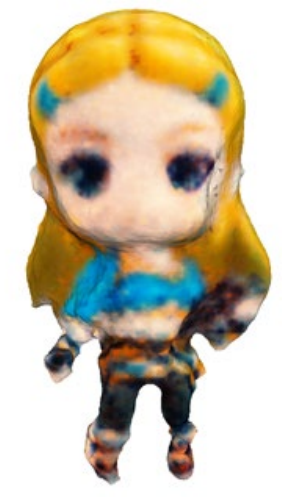} &
\includegraphics[width=0.115\textwidth]{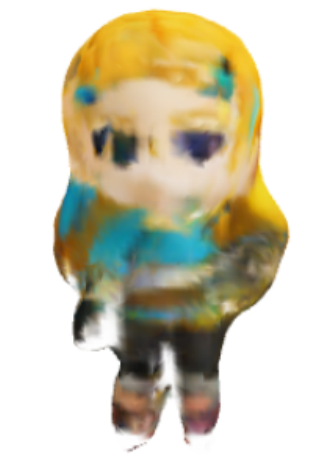} &
\includegraphics[width=0.096\textwidth]{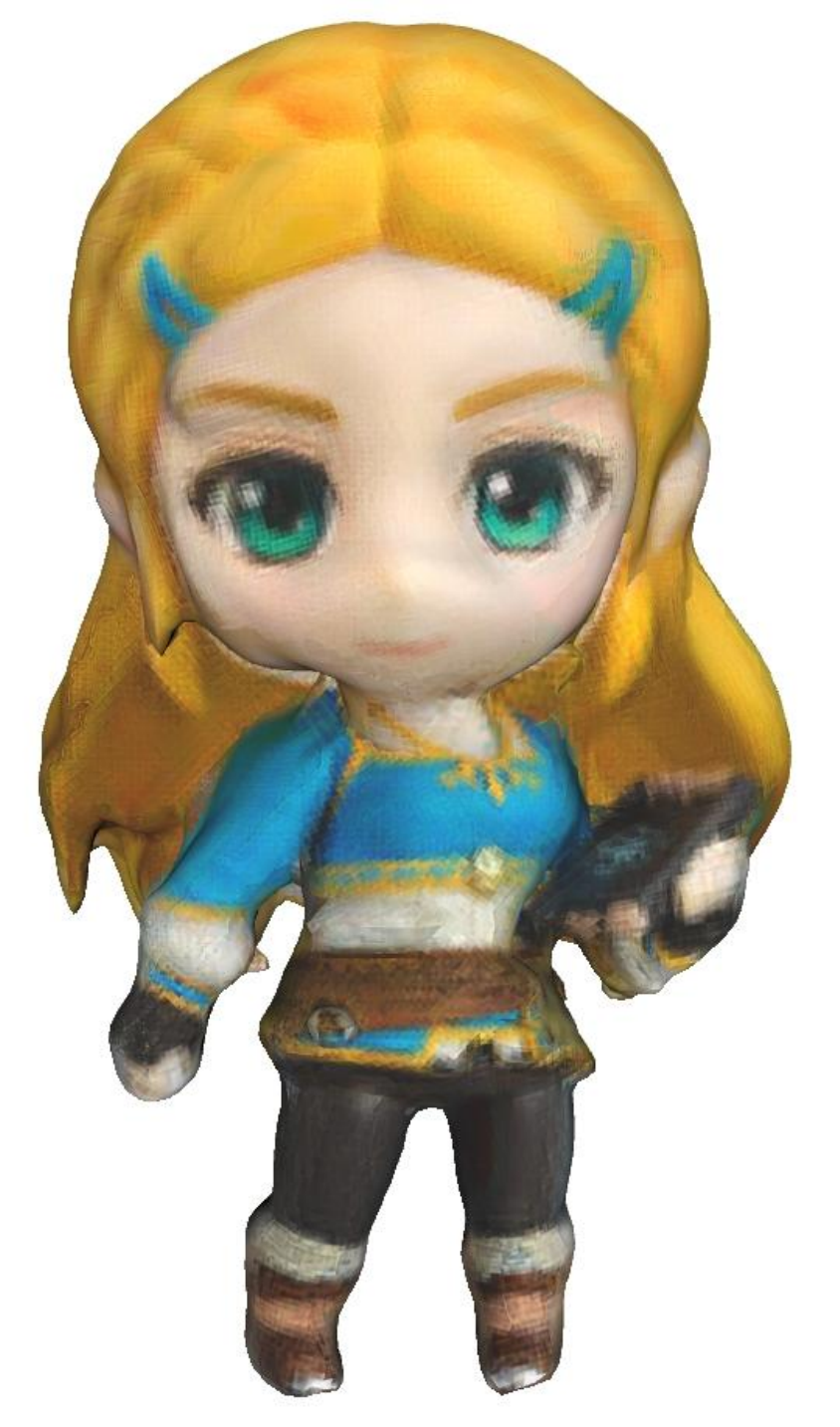} &
\includegraphics[width=0.11\textwidth]{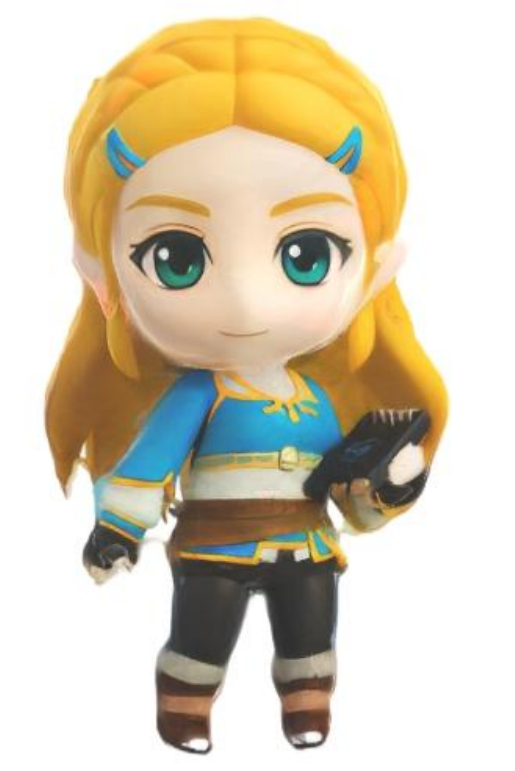} &
\includegraphics[width=0.089\textwidth]{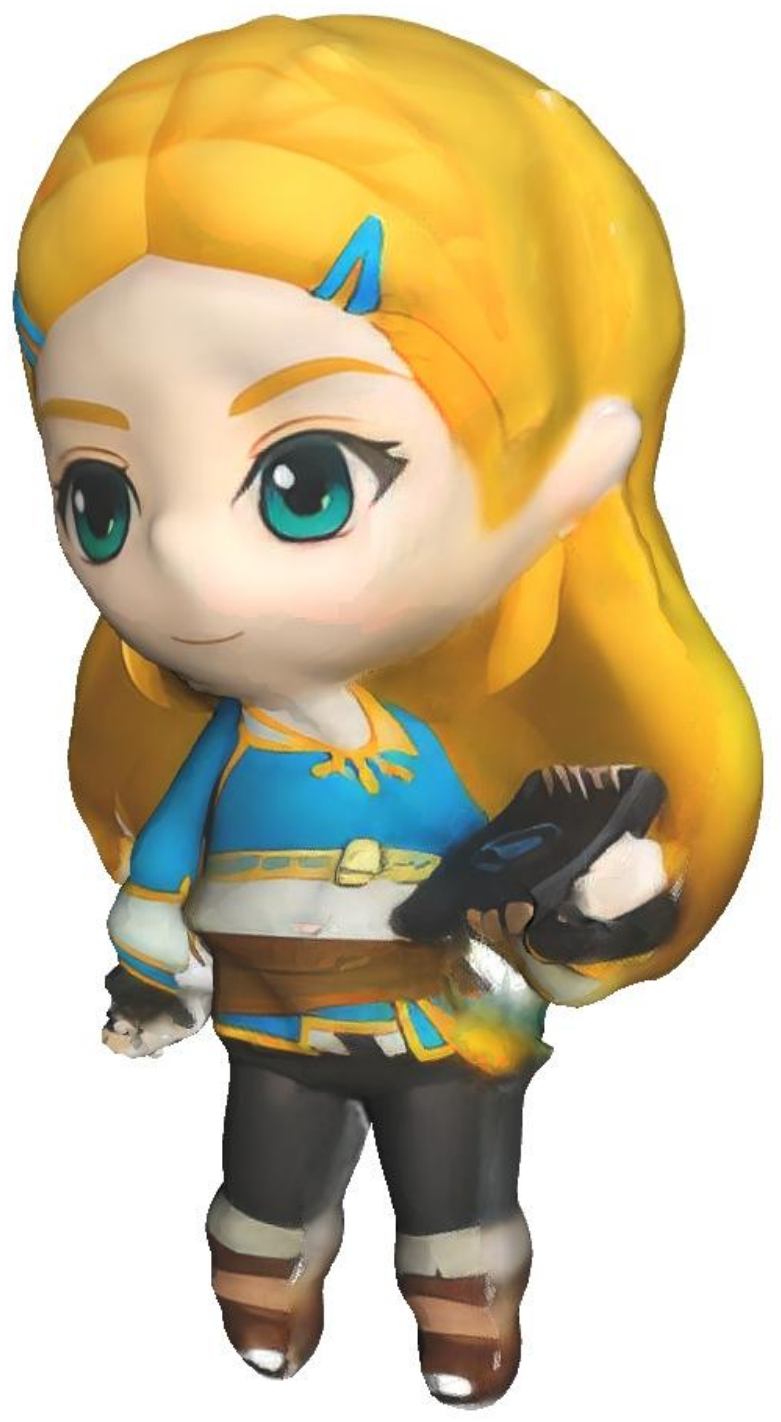} &
\includegraphics[width=0.11\textwidth]{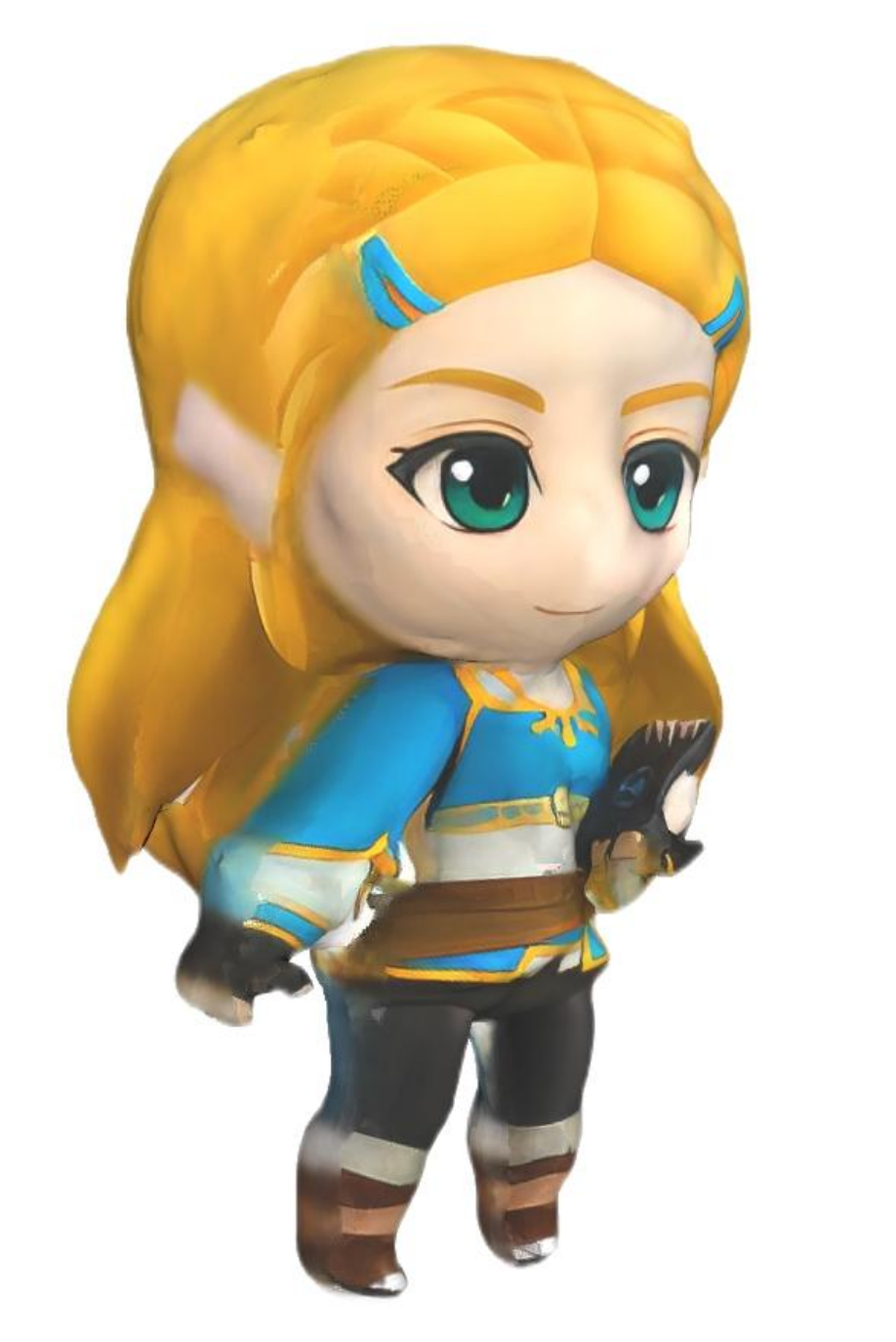} &
\includegraphics[width=0.11\textwidth]{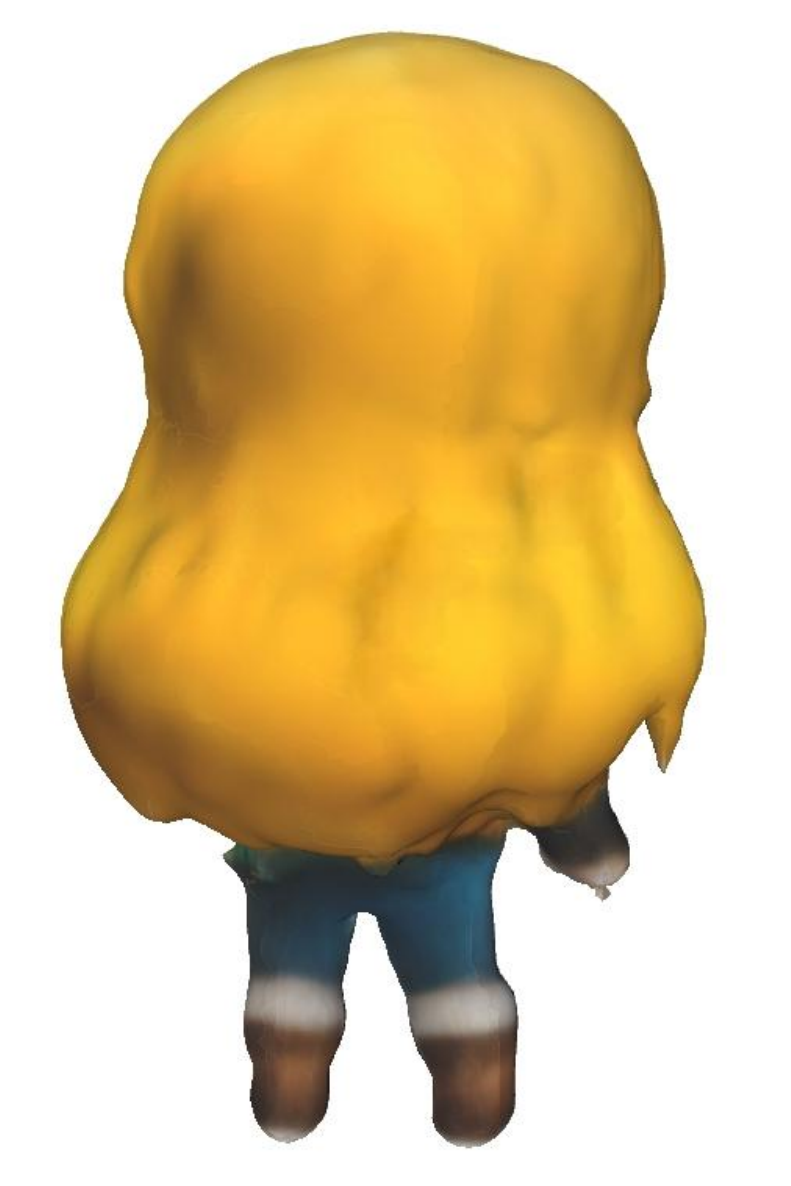} \\[-2pt]

\includegraphics[width=0.11\textwidth]{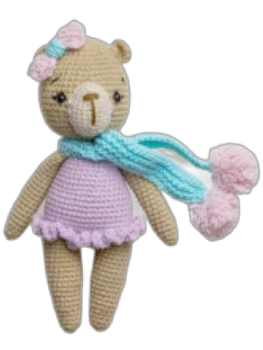} &
\includegraphics[width=0.12\textwidth]{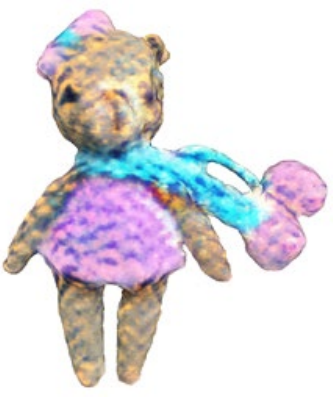} &
\includegraphics[width=0.12\textwidth]{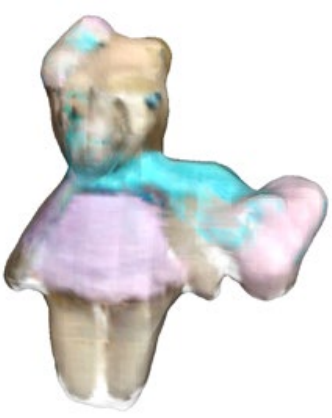} &
\includegraphics[width=0.12\textwidth]{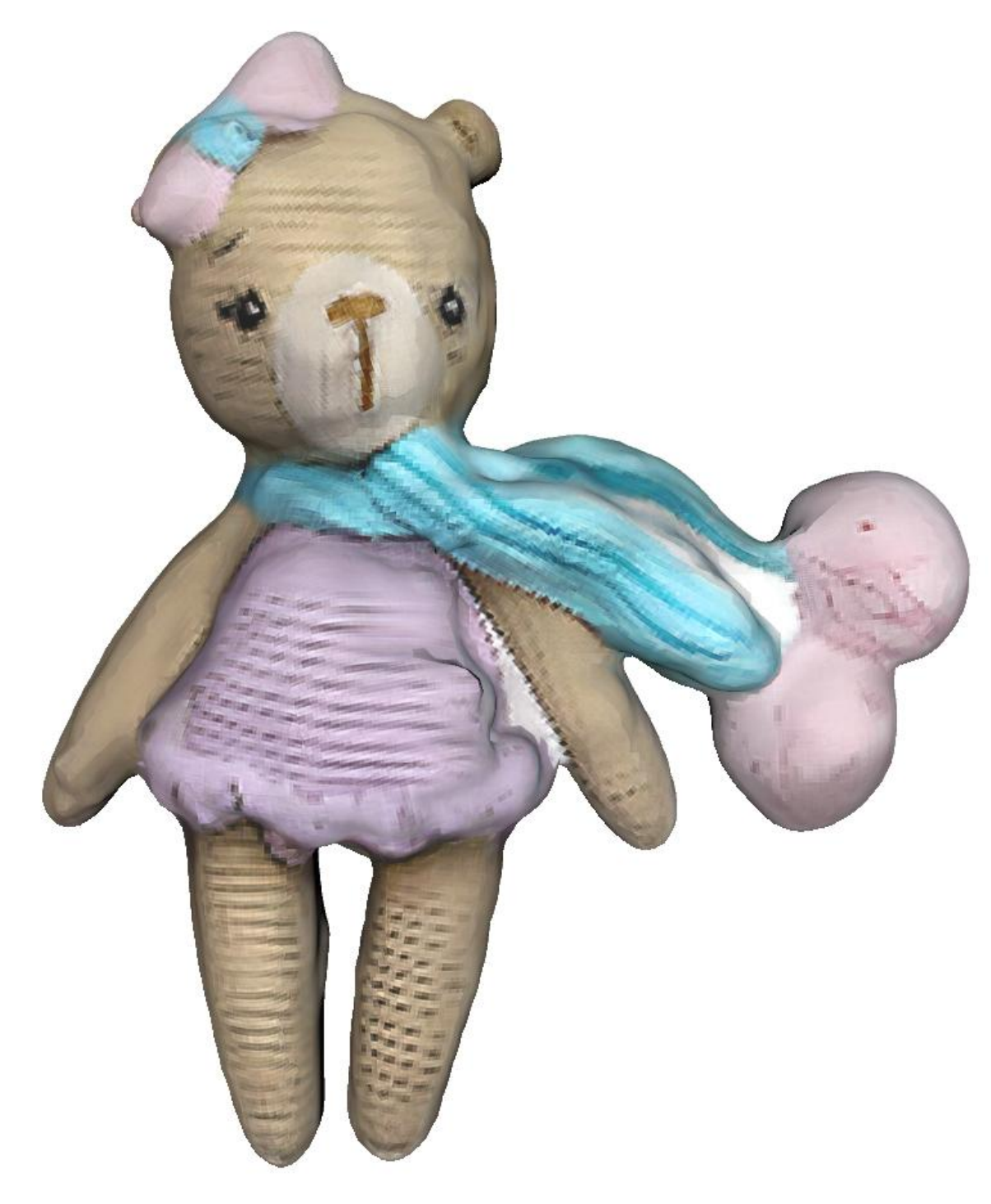} &
\includegraphics[width=0.12\textwidth]{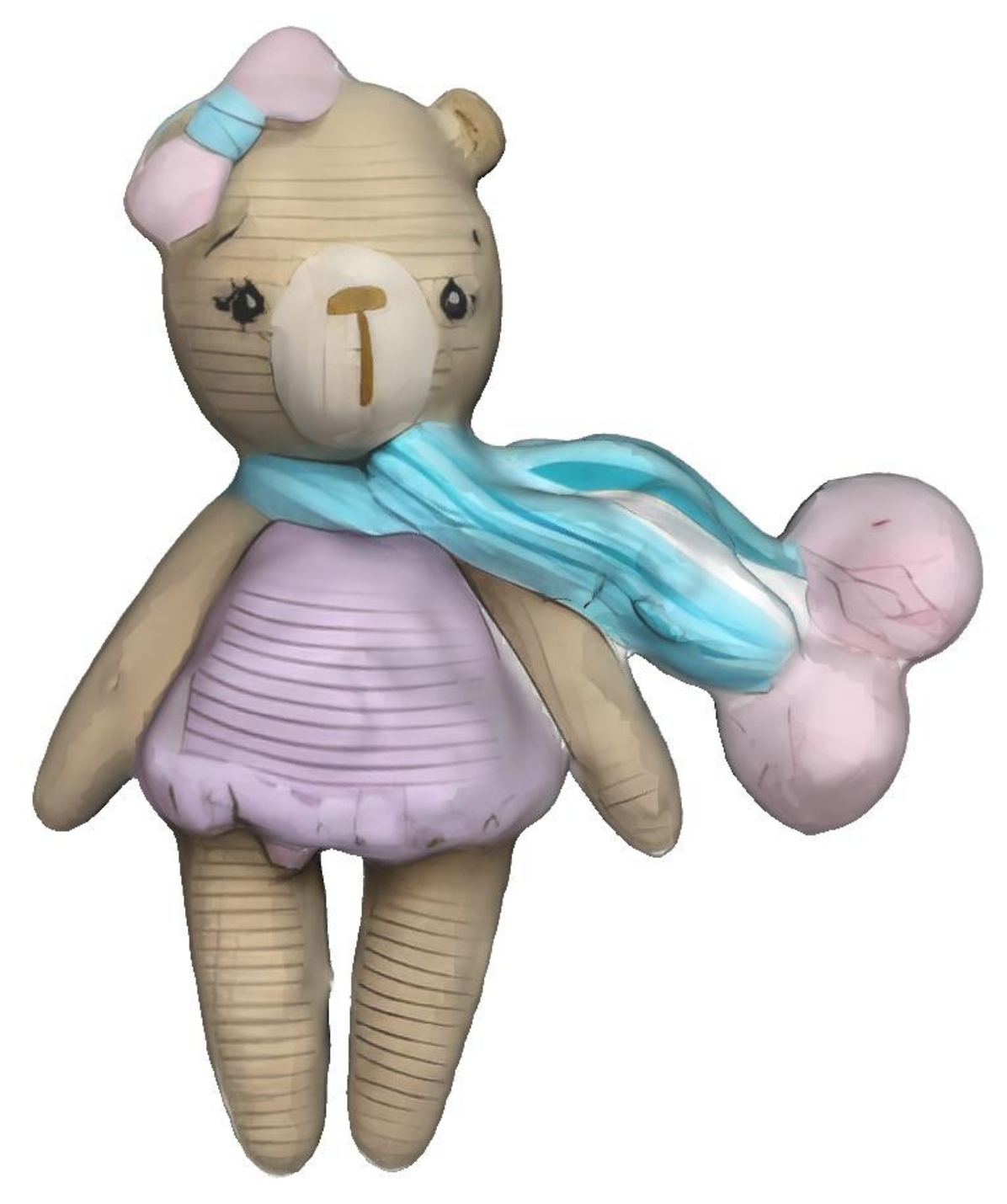} &
\includegraphics[width=0.113\textwidth]{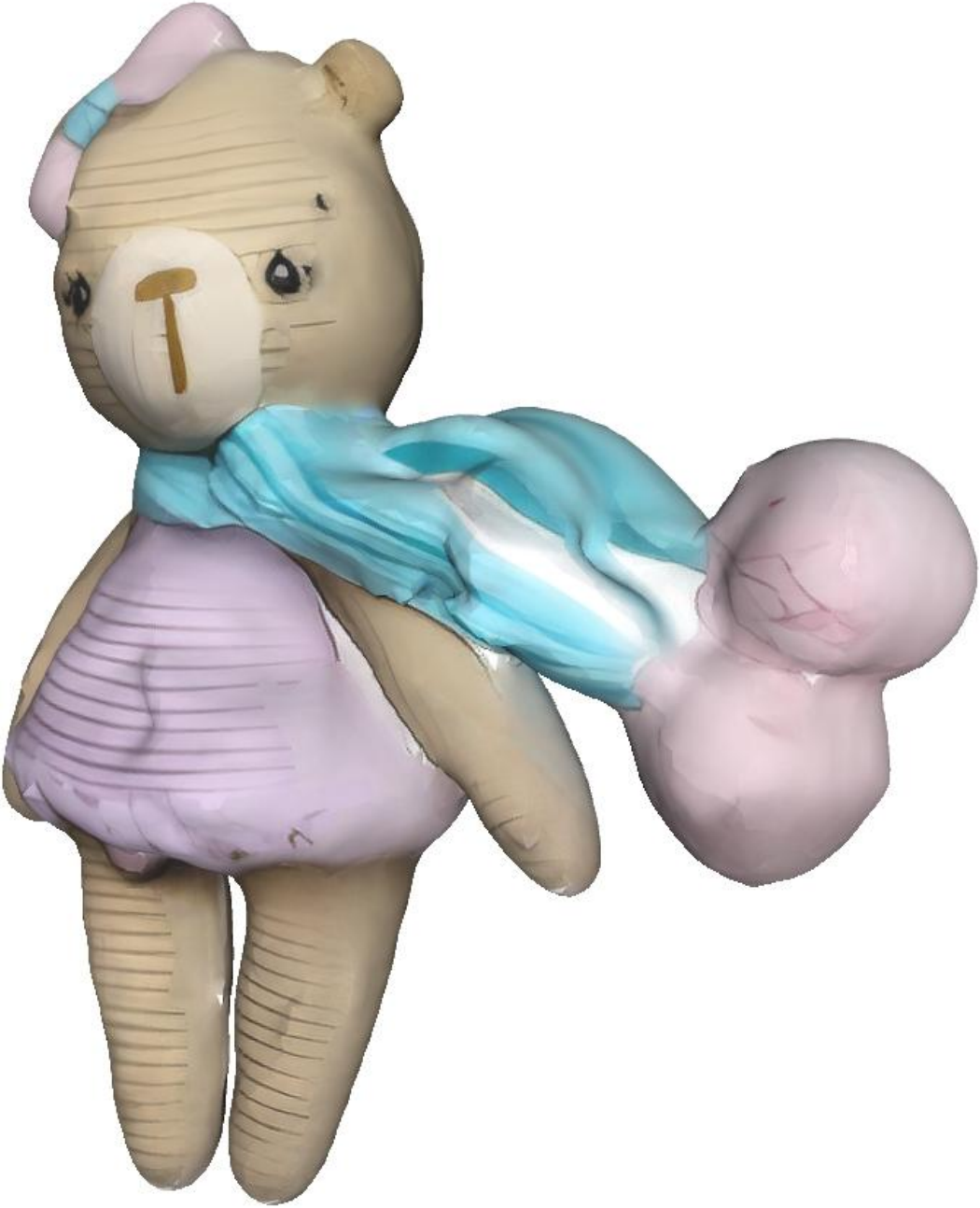} &
\includegraphics[width=0.10\textwidth]{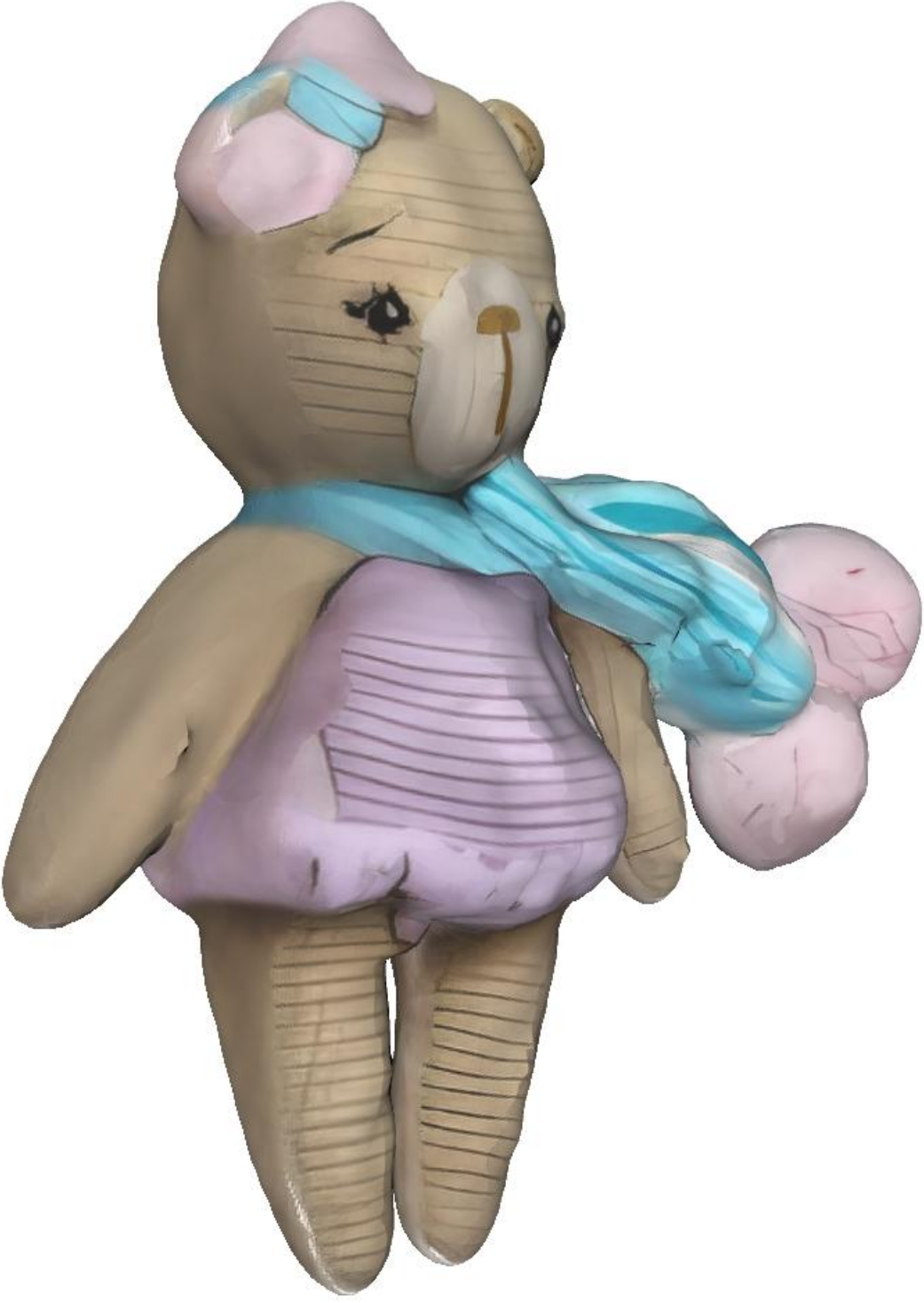} &
\includegraphics[width=0.135\textwidth]{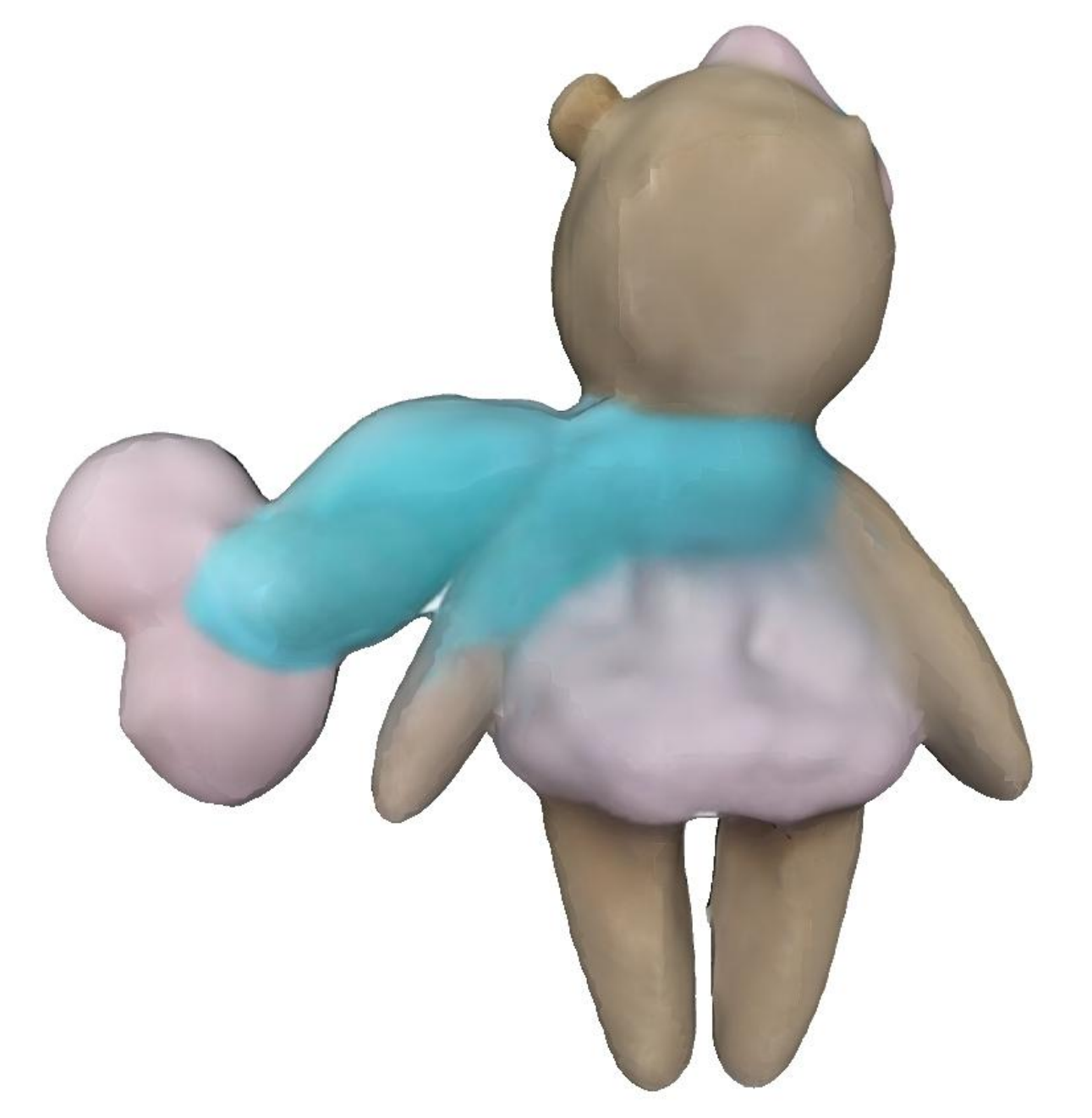} \\ [2pt]

Avg. Time(s)&
1200s &
45s &
60s &
\multicolumn{4}{c}{80s}\\[5pt]

Input &
Zero-1-to-3 &
One-2-3-45 &
DreamGaussian &
\multicolumn{4}{c}{Ours}\\
\end{tabular}
}
\caption{\textbf{Qualitative comparison.} We compare our method with Zero-1-to-3~\cite{22}, One-2-3-45~\cite{23}, and DreamGaussian~\cite{10}. The results show that our method provides superior visual quality and relatively faster generation speed.}
\label{fig:comparison_7col}
\end{figure*}

\subsection{Efficient Mesh Extraction}
To extract mesh geometry, it is essential to construct a dense density grid on which the Marching Cubes algorithm~\cite{27} can be applied. An important feature of the Gaussian splatting method is that it can dynamically split or prune oversized Gaussian distributions during the optimization process. Therefore, we utilize local density queries and color back-projection to fully leverage this property.

\textbf{Local Density Query.} Following the approach described by Tang et al.~\cite{11}, we divide the 3D space $(1,-1)^3$ into $16^3$ overlapping blocks and cull the Gaussian distribution lying outside each block to reduce the number of queries. We then query an $8^3$ dense grid for each block, resulting in a  $128^3$ final grid. At each grid position, we sum the weighted opacity of the remaining Gaussians:
\begin{equation}
d(\mathbf{x}) 
= \sum_{i=1}^K 
    \alpha_i 
    \exp\!\Bigl( 
      -\tfrac{1}{2}\,(\mathbf{x} - \mathbf{x}_i)^\top\,\Sigma_i^{-1}\,(\mathbf{x} - \mathbf{x}_i) 
    \Bigr),
\label{eq:density}
\end{equation}
where \(\alpha_i\) is the mixing coefficient for the \(i\)-th component, satisfying \(\sum_{i=1}^{K} \alpha_i = 1\). \(\mathbf{x}_i\) is the mean vector of the \(i\)-th Gaussian component. \(\Sigma_i\) is the covariance matrix of the \(i\)-th Gaussian constructed from scaling $s_i$ and rotation $r_i$. After calculating the density, we then extract the mesh surface using the Marching Cubes algorithm.

\textbf{Color Back-projection.}After obtaining the mesh geometry, we generate a baked texture by back-projecting rendered RGB images onto the mesh surface. Following UV unwrapping\cite{10.1145/3596711.3596734}, RGB images are rendered from eight azimuthal and three elevational angles, plus top and bottom views. Pixels are back-projected based on their UV coordinates, excluding those with low camera-space z-direction normals to avoid boundary instability~\cite{28}. This step produces an initial texture for further refinement.

\subsection{Texture Refinement}
In the second stage, we focus on refining the extracted rough textures to achieve a level of detail that closely matches the richness of the input image. Using the existing texture, we render a coarse image $\mathbf{I}^{p}_{coarse}$ from an random camera position $p$. This image is then perturbed with random noise and refined through a multi-step denoising process $f_{\phi} (\cdot)$ based on a 2D diffusion model:
\begin{equation}
I_{\text{refine}}^p = f_{\phi} \bigl( I_{\text{coarse}}^p + \epsilon(t_{\text{start}}); \, t_{\text{start}},\, c \bigr),
\end{equation}
where $\epsilon(t_{\mathit{start}})$ represents random noise at timestep $t_{\mathit{start}}$ and $c$ denotes the camera position $\Delta p$. The refined image is subsequently used to optimize the texture with an MSE loss:
\begin{equation}
L_{\mathit{MSE}} = \bigl\|I^p_{\mathit{refine}} - I^p_{\mathit{coarse}}\bigr\|_{2}^{2}.
\label{eq:MSE}
\end{equation}

\begin{table*}[t]
\centering
\caption{\textbf{Quantitative Comparison.} Zero-1-to-3* is an improved version of Zero-1-to-3 that incorporates mesh fine-tuning. A lower LPIPS value suggests better perceptual image quality. A higher CLIP-Similarity indicates greater alignment with the target.}
\resizebox{0.8\textwidth}{!}{
\begin{tabular}{ccccc}
\hline
\textbf{Methods} & \textbf{Type} & \textbf{LPIPS}$\downarrow$ & \textbf{CLIP-Similarity}$\uparrow$ & \textbf{Generation Time}$\downarrow$ \\
\hline
Shap-E~\cite{31} & Inference-only & 0.590 & 0.496 & \textbf{27s} \\
One-2-3-45~\cite{23} & Inference-only & 0.565 & 0.591 & 45s \\
Zero-1-to-3~\cite{22} & Optimization-based & 0.543 & 0.563 & 1200s\\
Zero-1-to-3*~\cite{22} & Optimization-based & 0.379 & 0.738 & 1800s \\
DreamGaussian~\cite{10} & Optimization-based & 0.389 & 0.653 & 60s \\
Ours & Optimization-based & \textbf{0.307} & \textbf{0.764} & 80s \\
\hline
\end{tabular}
}
\label{tab:quantitative_comparison}
\end{table*}

\begin{figure*}[!ht]
\centering
\includegraphics[width=0.8\textwidth]{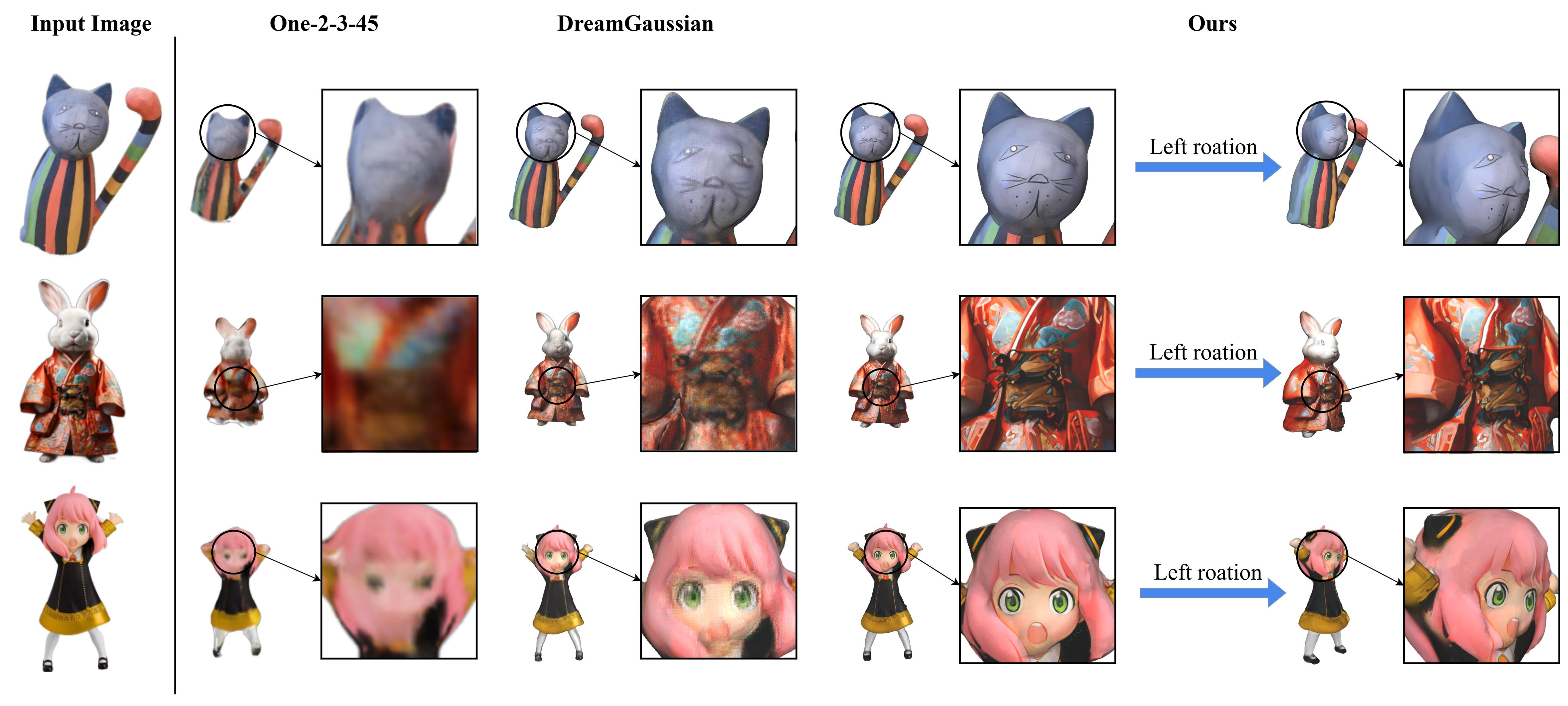}
\caption{\textbf{Detailed comparison.} We examine the finer details of the generated model, then rotate it to the left to check for any distortions.}
\label{fig:detailed_comparison}
\end{figure*}

\section{EXPERIMENTS}
\subsection{Implementation Details}
Our pipeline begins by preprocessing the input image through background removal~\cite{29}, recentering the foreground object, and resizing it to $256 \times 256$ resolution. We employ Real-ESRGAN~\cite{26} for super-resolution. Novel view image generation is conducted with a batch size of 2. 

We initialize 3D Gaussians with an opacity of 0.1 and grey color, uniformly distributed within a sphere of radius 0.5. Using 5000 random particles, we densify every 100 iterations to enhance density representation. The resolution for Gaussian splatting progressively increases from $64$ to $512$, while mesh rendering resolution is randomly sampled between $512$ and $1024$. 

For contrastive learning, we set an LPIPS threshold of 0.3, with a dynamically adjusted Quantity-Aware Triplet Loss margin for balanced training. Camera poses are sampled randomly at a fixed radius of 2, with a y-axis FOV of $49.1^\circ$, matching the Zero-1-to-3 setup. Azimuth angles range from $-180^\circ$ to $180^\circ$, and elevation angles span $-30^\circ$ to $30^\circ$. The weights for RGB and transparency loss linearly increased from 0 to 10$^{4}$ and 10$^{3}$, respectively. Mesh extraction is performed using the Marching Cubes algorithm with a density threshold of 1 for high-fidelity geometry. Stage~1 consists of 500 training steps, followed by 50 steps in Stage~2. All experiments run on an NVIDIA RTX 4090 (24\,GB) GPU, with our method using under 10\,GB of GPU memory.

\subsection{Qualitative and Quantitative Comparison}
Fig.~\ref{fig:comparison_7col} provides a qualitative comparison between our method and three advanced baselines: Zero-1-to-3~\cite{22}, One-2-3-45~\cite{23}, and DreamGaussian, using samples commonly referenced in Image-to-3D research. To provide a more intuitive comparison, we also highlight texture and geometry details in Fig.~\ref{fig:detailed_comparison}, showing accuracy in preserving subtle textures and geometric features. The visual results generated by our proposed models significantly outperform existing methods in terms of geometric precision and texture fidelity, underscoring the superiority of our approach in producing high-resolution outputs with intricate structural details.

Table~\ref{tab:quantitative_comparison} presents a quantitative comparison of multiple Image-to-3D methods using LPIPS~\cite{25}, CLIP-similarity~\cite{30} and average generation time. For the assessment, we selected images from previous works. As shown, our contrastive-learning-based approach significantly outperforms existing methods in generating high-fidelity 3D contents, while maintaining competitive efficiency. Although slightly slower than DreamGaussian, our method still operates within a comparable time frame and substantially accelerates the generation process relative to other optimization-based methods. These findings indicate that our method balances quality and generation speed, offering performance close to optimization-based techniques with only a slight processing overhead versus inference-only methods. 
\begin{figure}[t]
    \centering
    \begin{subfigure}[t]{0.40\columnwidth}
        \centering
        \includegraphics[width=\textwidth]{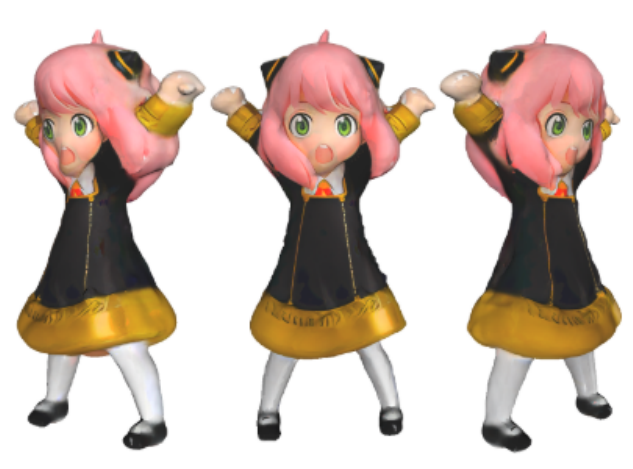}
        \caption*{Full setting}
    \end{subfigure}\qquad
    \begin{subfigure}[t]{0.40\columnwidth}
        \centering
        \includegraphics[width=\textwidth]{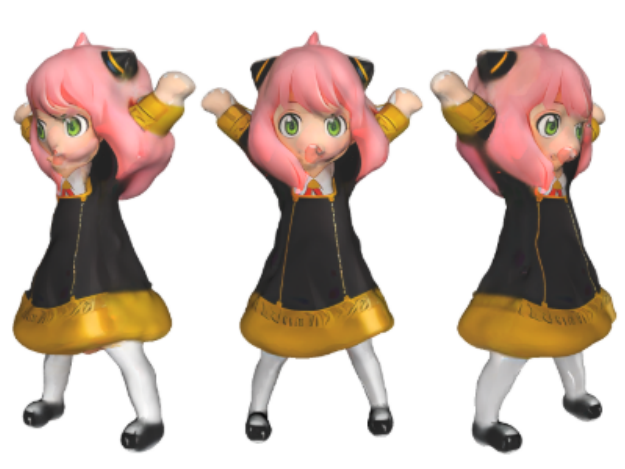}
        \caption*{w/o Super-resolution}
    \end{subfigure}
    
    \vspace{0.1cm} 
    
    \begin{subfigure}[t]{0.40\columnwidth}
        \centering
        \includegraphics[width=\textwidth]{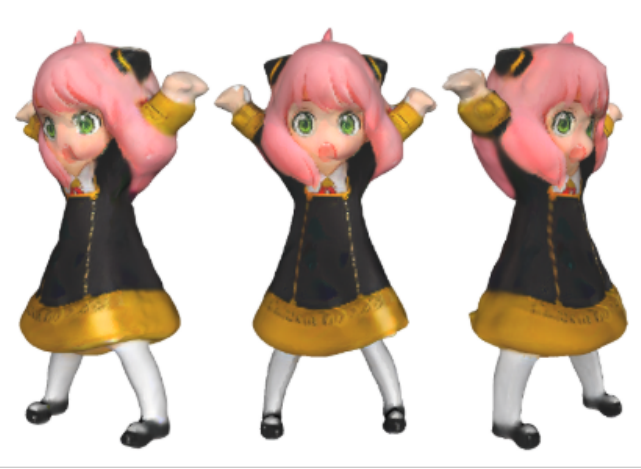}
        \caption*{w/o QA-Triplet loss}
    \end{subfigure}\qquad
    \begin{subfigure}[t]{0.43\columnwidth}
        \centering
        \includegraphics[width=\textwidth]{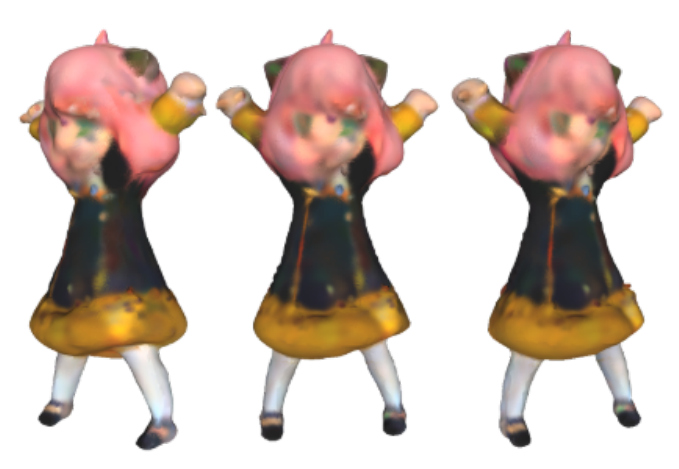}
        \caption*{w/o Texture refinement}
    \end{subfigure}
    
    \caption{
        \textbf{Ablation Study.} 
        We ablate the proposed designs in our framework to verify their effectiveness.
    }
    \label{fig:ablation}
\end{figure}

\subsection{Ablation Study}
We perform ablation studies on key design components of our method, as shown in Fig.~\ref{fig:ablation} and Table~\ref{tab:ablation}. Our analysis primarily focuses on the significance of the Super-resolution technique and the QA-Triplet Loss, given the previous mesh extraction and texture refinement method. Our findings show that omitting either of these components degrades the quality of the generated models. Specifically, while the super-resolution technique substantially enhances texture details, it may also introduce geometric distortions and texture artifacts. In contrast, the QA-Triplet Loss is essential for improving geometric fidelity while preserving texture quality, thereby mitigating these potential distortions. This underscores the complementary roles of both techniques in achieving high-quality 3D content generation. Furthermore, texture refinement, the core design of the second phase, further enhances texture quality and is also essential to our framework.

\begin{table}[t]
    \centering
    \caption{\textbf{Quantitative Comparison for Ablation Study.}}
    \begin{tabular}{cccc}
        \toprule
        Settings & LPIPS & CLIP-Similarity \\
        \midrule
        w/o Super-resolution & 0.343 & 0.691 \\
        w/o QA-Triplet Loss & 0.337 & 0.708 \\
        w/o Texture refinement & 0.404 & 0.573 \\
        Full setting & \textbf{0.307} & \textbf{0.764} \\
        \bottomrule
    \end{tabular}
    \label{tab:ablation}
\end{table}

\section{CONCLUSION}
In this work, we introduce ContrastiveGaussian, a high-fidelity 3D model generation framework that significantly enhances both the efficiency and quality of 3D model creation. By integrating high-resolution contrastive learning with advanced 2D diffusion models into the Gaussian splatting pipeline, our approach effectively balances visual detail and computational efficiency. Additionally, we propose a novel Quantity-Aware Triplet Loss, which dynamically adapts to varying sample distributions. Coupled with a dedicated texture refinement stage, our framework significantly lowers computation time relative to optimization-based methods while producing high-quality mesh geometry and detailed textures from a single image. Future work will focus on refining object geometry and texture, particularly from less frequently observed perspectives, such as rear views.

\bibliographystyle{IEEEtran}
\bibliography{reference}

\end{document}